\documentclass[preprint,11pt,3p,times]{elsarticle}

\usepackage{amsmath,amsthm,amssymb,amsfonts,bbm}
\usepackage{multirow,multicol}
\usepackage{comment}
\usepackage{booktabs}
\usepackage[ruled]{algorithm}
\usepackage{algorithmic}
\usepackage{soul}
\usepackage{xcolor}
\usepackage{caption}
\usepackage{subfigure,subcaption}
\usepackage{wrapfig}
\usepackage{bm}
\usepackage{ulem}

\usepackage[symbol*]{footmisc}
\setfnsymbol{wiley}

 \bibpunct[, ]{(}{)}{,}{a}{}{,}%

\def\bff{\mathbf{f}}

\def\bK{\mathbf{K}}

\def\bTheta{{\boldsymbol\Theta}}

\def\bSigma{{\boldsymbol\Sigma}}

\def\bPsi{{\boldsymbol\Psi}}
\def\bOmega{{\boldsymbol\Omega}}

\def\bbf{\mathbf{f}}

\def\bt{\mathbf{t}}
\def\bu{\mathbf{u}}

\def\by{\mathbf{y}}
\def\bz{\mathbf{z}}

\def\btheta{{\boldsymbol\theta}}

\def\bmu{{\boldsymbol\mu}}

\def\bxi{{\boldsymbol\xi}}

\def\bphi{{\boldsymbol\phi}}

\def\bbE{\mathbb{E}}

\def\bbR{\mathbb{R}}

\def\cD{\mathcal{D}}

\def\cL{\mathcal{L}}

\def\cN{\mathcal{N}}

\def\cV{\mathcal{V}}

\def\KL{{\sf KL}}

\def\ones{\mathbf{1}}

\def\zeros{\mathbf{0}}

\journal{Journal}

\begin{document}

\begin{frontmatter}

\title{\textbf{Fed-Joint}: Joint Modeling of Nonlinear Degradation Signals and Failure Events for Remaining Useful Life Prediction using Federated Learning}

\author[label1]{Cheoljoon Jeong} 
\affiliation[label1]{organization={University of Michigan},
            addressline={1205 Beal Avenue}, 
            city={Ann Arbor},
            postcode={48105}, 
            state={MI},
            country={USA}}
\author[label2]{Xubo Yue}
\affiliation[label2]{organization={Northeastern University},
            addressline={110 Forsyth Street}, 
            city={Boston},
            postcode={02115}, 
            state={MA},
            country={USA}}
\author[label3]{Seokhyun Chung\footnote{Corresponding author.}}
\affiliation[label3]{organization={University of Virginia},
            addressline={151 Engineer's Way}, 
            city={Charlottesville},
            postcode={22903}, 
            state={VA},
            country={USA}}

\begin{abstract}
    Many failure mechanisms of machinery are closely related to the behavior of condition monitoring (CM) signals. To achieve a cost-effective preventive maintenance strategy, accurate remaining useful life (RUL) prediction based on the signals is of paramount importance. However, the CM signals are often recorded at different factories and production lines, with limited amounts of data. Unfortunately, these datasets have rarely been shared between the sites due to data confidentiality and ownership issues, a lack of computing and storage power, and high communication costs associated with data transfer between sites and a data center. Another challenge in real applications is that the CM signals are often not explicitly specified \textit{a priori}, meaning that existing methods, which often usually a parametric form, may not be applicable. To address these challenges, we propose a new prognostic framework for RUL prediction using the joint modeling of nonlinear degradation signals and time-to-failure data within a federated learning scheme. The proposed method constructs a nonparametric degradation model using a federated multi-output Gaussian process and then employs a federated survival model to predict failure times and probabilities for in-service machinery. The superiority of the proposed method over other alternatives is demonstrated through comprehensive simulation studies and a case study using turbofan engine degradation signal data that include run-to-failure events.
\end{abstract}




\end{frontmatter}

\section{Introduction}
\label{sec:intro}

Predicting the remaining useful life (RUL) of complex systems, such as machinery, is a pivotal component in maintenance planning and reliability engineering. Accurate RUL predictions enable operators to timely address potential failures, thereby reducing downtime, optimizing maintenance schedules, and minimizing costs associated with unexpected breakdowns or over-maintenance \citep{ferreira2022remaining}. 

Among the various approaches for RUL prediction, data-driven methods leverage machine learning and statistical models to analyze historical and real-time condition monitoring (CM) data (e.g., sensor readings), identifying correlations between system degradation and RUL \citep[e.g.,][]{ma2020deep, chen2020machine, mo2021remaining, rauf2022machine, guo2023deep, keshun2024optimizing}. These methods excel at handling large high-dimensional datasets and can adapt to various failure modes without requiring explicit knowledge of the underlying physical mechanisms. This makes data-driven approaches particularly valuable compared to physics-based methods, especially in complex systems where detailed knowledge of material properties and failure mechanisms is unavailable or challenging to model.

\textbf{Joint modeling} approaches have recently gained significant traction in data-driven RUL prediction due to their ability to integrate degradation data with survival analysis \citep[e.g.,][]{zhou2014remaining, rizopoulos2017dynamic}. They respectively regard a system's remaining life and failure time as the remaining survival time and time-to-event data. Joint modeling captures the dynamic relationship between degradation patterns and failure risk over time. Doing so provides several key advantages inherited from survival analysis, including the capability to dynamically incorporate time-varying covariates, enhanced interpretability of predictors, and probabilistic RUL predictions. This integrated framework is particularly suitable for applications where degradation patterns are closely linked to failure probabilities.

Despite the key advantages above, existing joint modeling methods for RUL prediction typically rely on centralized data for analysis. These methods aggregate CM data and failure time records from all historical assets (or units) into a central repository. Building upon the aggregated data, centralized approaches facilitate the training of machine learning or statistical models that can capture the temporal evolution of CM data and its relationship with failure events. However, such a framework faces considerable challenges, especially in scenarios where data is collected from multiple distributed sources (or sites).

For instance, in the airline industry, individual companies maintain proprietary data for their fleet components, such as aircraft engines. Each company can build predictive models using CM data and historical maintenance records to predict RUL for its own fleet. While a more robust RUL prediction model could be achieved by training on a larger centralized dataset aggregated from multiple companies, this approach raises several critical issues, such as privacy concerns, data ownership conflicts, and logistical challenges associated with aggregating vast amounts of data from geographically dispersed sources---issues that represent significant barriers.

Moreover, independent modeling within each company may not be desired. Local datasets are often small and unrepresentative of the broader population of assets, leading to models that lack generalization and fail to deliver accurate RUL predictions across diverse operational conditions. This challenge is further exacerbated when dealing with complex systems, which produce highly nonlinear CM data requiring a substantial volume of samples to model effectively. Therefore, building alternative joint modeling approaches that enable collaborative RUL prediction while addressing privacy concerns is necessitated.

\begin{figure}[t!]
    \centering
\includegraphics[width=0.8\linewidth]{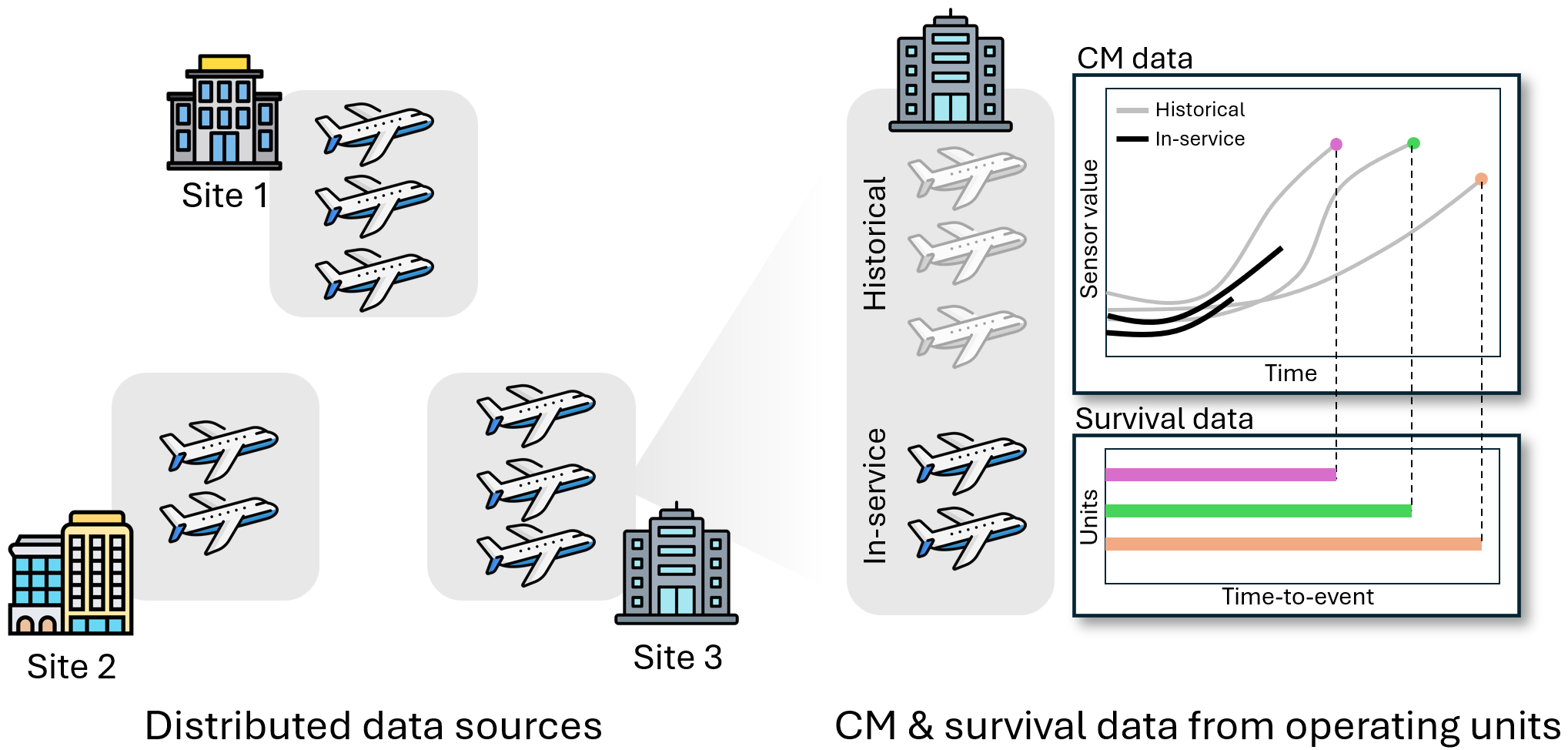}
    \caption{Distributed CM and survival data.}
    \label{fig:dist}
\end{figure}

To this end, \textbf{this study presents new federated joint modeling, referred to as \textit{Fed-Joint}}, a collaborative framework to predict the RUL of assets  distributed across sites and estimate their failure probability within a specific future time. Figure \ref{fig:dist} illustrates distributed sources for CM and survival data considered in this study. Our framework combines a multi-output Gaussian process (MGP) for nonparametric degradation modeling within a Cox proportional hazards (CoxPH) model for survival analysis in a federated scenario, where each site has its own CM and survival data obtained from their assets. The MGP component can capture the intricate and potentially nonlinear degradation patterns in the CM signals generated by assets over time. Estimating cross-unit correlation based on their signals enables the use of MGP modeling in a nonparametric way. Meanwhile, the CoxPH model component is designed to discern the relationship between estimated CM patterns and time-to-event data, which is crucial for understanding and predicting failure events for units. Importantly, both modeling schemes are jointly analyzed  without sharing the raw CM and survival data across sites. This federated joint model operates through iterations of two steps: (i) each local site trains its own joint model on its local data; and (ii) only the necessary model parameters are exchanged across the sites, ensuring that raw data remains localized and secure. This collaborative yet decentralized learning process not only addresses data privacy concerns but also leverages the diverse data sources to improve the overall model accuracy, robustness, and generalizability.

\textbf{Our key innovations} include the integration of the MGP and CoxPH models within a federated learning (FL) framework. The nonparametric nature of the MGP provides flexibility to model complex behaviors of degradation signals without assuming a specific functional form, while the CoxPH model offers a robust statistical framework for analyzing failure events. By jointly modeling these two components in a federated setting, our approach delivers more reliable RUL predictions while mitigating the challenges associated with centralized data processing. \textbf{To the best of our knowledge, this is the first study to extend traditional centralized, parametric joint modeling into a federated, nonparametric joint modeling framework.} By introducing this federated joint modeling approach, our work contributes to the field of predictive maintenance by providing a scalable, privacy-preserving, yet collaborative solution for RUL prediction in reliability engineering. Comprehensive simulation and case studies demonstrate that our approach significantly improves prediction accuracy compared to independent joint modeling, while being competitive with a centralized approach that sacrifices confidentiality. Our work provides industries with a viable path to adopting collaborative predictive maintenance strategies without compromising data security.

The remainder of this paper is organized as follows. In Section~\ref{sec:literature}, we review related work on RUL prediction in the federated setting and joint modeling of longitudinal and time-to-event data. Section~\ref{sec:method} presents the formulation of our federated joint model, detailing the mathematical foundations of the MGP and CoxPH models and their integration. In Sections~\ref{sec:simulation} and \ref{sec:case}, we apply our proposed model to several synthetic and real datasets, demonstrating its effectiveness and comparing its performance with other alternatives. Finally, Section~\ref{sec:conclusion} concludes the article, highlighting the contributions and potential future research directions.

\section{Literature Review}
\label{sec:literature}

\paragraph{Federated RUL Prediction} In recent years, there has been a growing interest in predicting RUL using privacy-preserving FL techniques. Some notable work includes using deep learning techniques such as feedforward network \citep{rosero2020remaining, llasag2023forecasting, llasag2024label}, convolution network \citep{guo2022fedrul, chen2023federated2, chen2023remaining2, chen2023bearing1, luo2023enabling, cai2024fedcov}, transformer-based models \citep{du2023trans, kamei2023comparison, zhu2024collaborative}, adversarial network \citep{zhang2024multi}, long short-term memory  \citep{bemani2022aggregation, quan2022privacy, chen2023remaining, lv2024state, chung2023federated}, and autoencoder \citep{qin2023dynamic, wu2023privacy, zhong2024lithium, chen2024differential}. However, despite their success, these approaches typically yield \textbf{global models} that may not perform optimally across diverse and heterogeneous datasets. This limitation arises from the inherent assumption that all participants' data is similar, which is often not the case in real-world scenarios. Therefore, there is an increasing need for \textbf{personalized models} that can adapt to the unique characteristics of each data source, ensuring more accurate and reliable RUL predictions. Along this line, several efforts have aimed to tailor global models to better suit the personalized needs of individual clients. For example, \cite{arunan2023federated} introduced a feature similarity-matched parameter aggregation algorithm that selectively learns from heterogeneous edge data to provide personalized solutions for each client. Similarly, \cite{von2024data} proposed a data disparity-aware aggregation rule, ensuring that underrepresented clients in earlier rounds have a greater influence on later model updates. This approach mitigates bias and promotes equitable performance across clients. However, these previous efforts primarily focus on direct RUL prediction models and do not address survival analysis for time-to-event data.

\paragraph{Joint Modeling of longitudinal and time-to-event data} Our proposed Fed-Joint is related to the literature on a joint model of longitudinal and time-to-event data. \cite{rizopoulos2010jm,rizopoulos2011dynamic} employed generalized mixed-effects models to analyze longitudinal data and calculated the time-to-event distribution based on the mean predictions derived from the longitudinal model. This framework has been extensively applied to many applications in the literature~\citep{zhou2014remaining,proust2014joint,he2015simultaneous,rizopoulos2017dynamic,mauff2020joint}. In most joint models, the two-stage estimation procedure has been commonly employed because of the intractability of the joint likelihood and the significant computational complexity involved. The two-stage procedure estimate the underlying trend of the longitudinal data and this estimated curve is applied to the survival model to predict event probabilities. It is known that such an approach induces a bias. However, researchers pointed out that the two-stage procedure has a competitive prediction capability~\citep{wulfsohn1997joint,yu2004joint,zhou2014remaining,mauff2020joint}. In this study, we adopt the two-stage procedure to estimate the parameters of the proposed Fed-Joint model. This parameter estimation procedure will be discussed in the federated setting in Section~\ref{sec:inference}.

Also, many studies continue to rely on strict parametric assumptions, where it is presumed that signals adhere to a specific parametric shape and that all signals follow the same functional pattern. Essentially, signals are expected to exhibit similar behavior but at varying rates. However, parametric models can be quite limiting in many scenarios, and if the chosen form is significantly different from reality, the predictive outcomes may be inaccurate. Recent work by \citet{yue2021joint} introduced a joint modeling framework leveraging MGPs to handle nonparametric CM data. Their approach demonstrated superior performance compared to parametric methods, as MGPs effectively model nonparametric degradation curves. However, this framework relies on centralized analysis, whereas our study extends the nonparametric capabilities of MGPs into federated scenarios for joint modeling. We will explore the development of nonparametric models using MGPs in a federated manner in Section~\ref{sec:model_development}.

\section{Methodology}
\label{sec:method}

Suppose that there are $K$ sites or data sources (e.g., companies, factories, production lines, etc.) and each site $k$ for $k = 1,\ldots,K$, has $M_k$ units or assets (e.g., machines, tools, etc.). In each site $k$, a unit $m$ collects data $\cD_{k,m} = \{V_{k,m}, \delta_{k,m}, \bt_{k,m}, \by_{k,m}, \bm{w}_{k,m}\}$ for $m=1,\ldots,M_k$, where $V_{k,m} = \min(T_{k,m},C_{k,m})$ is the event time (i.e., the unit either failed at time $T_{k,m}$ or censored at time $C_{k,m}$) and $\delta_{k,m} \in \{0,1\}$ an event indicator, with $1$ indicating that the unit has failed or $0$ being censored. In this study, we consider only right-censoring without tied events for simplicity. However, this assumption can be easily relaxed within our framework. Additionally, for site $k$ and unit $m$, $\bt_{k,m}=[t_{k,m,1},\ldots,t_{k,m,L_{k,m}}]^T$ is the ordered timestamps, $\by_{k,m} = [ y_{k,m}(t_{k,m,1}), \ldots, y_{k,m}(t_{k,m,L_{k,m}}) ]^T$ is the observations measured at $\bt_{k,m}$ (e.g., machine's observed degradation signal) with $t_{k,m,l} \le V_{k,m}$ for $l=1,\ldots,L_{k,m}$, and $\bm{w}_{k,m}$ is a vector of covariates that contains all time-invariant information about the corresponding unit (e.g., machine's type). Note that the interval between two consecutive timestamps in $\bt_{k,m}$ can be either regular or irregular, depending on the problem under study. 
For the sake of brevity, we also define collective notations: $\bt = [\bt^T_{k,m}]^T_{k=1,\ldots,K,\;  m=1,\ldots,M_k}\in \bbR^{ L \times 1}$ and $\by = [\by^T_{k,m}]^T_{k=1,\ldots,K,\;  m=1,\ldots,M_k} \in \bbR^{ L \times 1} $, where $L = \sum_{k=1}^{K} \sum_{m=1}^{M_k} L_{k,m}$ and $\sum_{k=1}^{K} M_k = M$.

Suppose that either data sharing across sites or aggregating all the data in a central repository is not possible, except for sharing the model or its parameters. Using the data $\cD_{k,m}$ for all $k=1,\ldots,K$ and $m=1,\ldots,M_k$, we propose a new prognostic framework that employs a joint model of degradation signals and time-to-failure data using federated learning. We utilize a two-stage estimation procedure with federated inference for model parameters by introducing the concepts from longitudinal and survival analysis subsequently. 

\subsection{Model Development}
\label{sec:model_development}

\subsubsection{Nonparametric modeling and extrapolation of time-dependent observations}
\label{sec:fed_mgp}
The first step of our approach is extrapolating the future trajectory of time-dependent observations $\by_{k,m}$ for each unit $m$ at site $k$. 
The key idea is to extract and utilize shared information across all units at different sites. This strategy allows for accurate extrapolation at the sites where most units are either in the early stages of degradation or have sparse observations, by leveraging data from other sites that have dense observations by monitoring their units throughout entire degradation cycles. Yet, here lie two challenges. First, time-dependent observations are often nonlinear. This makes parametric modeling prone to model misspecification and poor curve estimation. Second, building a method that leverages information across units from different sites usually requires centralized data processing, that is, collecting all data across sites into a central location for model development. Such a framework raises substantial concerns about deteriorated privacy and necessitates significant computational and storage resources for handling data along with data transfer costs.

We consider a method that can address the critical issues above simultaneously. Specifically, we build upon federated multi-output Gaussian processes (FedMGPs), proposed by \cite{chung2024federated}, an MGP model designed for federated scenarios. This approach allows for nonparametric modeling of time-dependent observations and, at the same time, accurate extrapolation through knowledge sharing across sites without the need for local data sharing nor centralized computation.

Suppose a regression function $f_{k,m}: \bbR_+ \rightarrow \bbR$ that relates an arbitrary input $t$ to an output $y_{k,m}(t)$ for unit $m$ at site $k$, represented as
\begin{equation}\label{eq:func}
    y_{k,m}(t) = f_{k,m}(t) + \epsilon_{k,m}(t),
\end{equation}
where $\epsilon_{k,m}(t) \sim \cN(0, \sigma_{k,m}^{2})$ is a Gaussian noise. The core idea to enable sharing knowledge across sites is to establish correlations among $f_{k,m}(t)$ across all $k=1,\ldots,K$, rather than modeling either each $f_{k,m}(t)$ independently or $f_{k,m}(t)$'s within a respective site $k$ only. One way to do so is based on convolution process (CP)-based MGP construction~\citep{alvarez2008sparse}. Consider a set of independent latent functions $u_i(t)$ for $i = 1,\ldots,I$ that are shared across all units. Then, each function $f_{k,m}(t)$ is modeled as a summation of $I$ functions, with each obtained by convolving the shared latent function $u_i(t)$ using a smoothing kernel $g_{k,m,i}(\cdot; \btheta_{k,m,i})$ with the parameter $\btheta_{k,m,i}$, for $i=1,\ldots,I$. This is written as
\begin{equation*} 
  f_{k,m}(t) = \sum_{i=1}^I\int^\infty_{-\infty} g_{k,m,i}(t - \tilde t; \btheta_{k,m,i})u_i(\tilde t)d\tilde t.  
\end{equation*}
If we model the latent functions $u_1(t),\ldots,u_I(t)$ by independent Gaussian processes (GPs) with kernels $g_{u_i}(\cdot, \cdot; \bxi_i)$ for the parameter $\bxi_i$, the covariance between the values of the function $f_{k,m}(t)$ and the latent function $u_i(t')$ is 
\begin{equation}\label{eq:cov_fu}
    \operatorname{cov}(f_{k,m}(t), u_{i}(t')) =  \int_{-\infty}^{\infty} g_{k,m,i}(t - \tilde t; \btheta_{k,m,i})g_{u_i}(t', \tilde t; \bxi_i)d\tilde t,  
\end{equation} and the covariance between the values of the functions $f_{k,m}(t)$ and $f_{k',m'}(t')$ is 
\begin{equation}\label{eq:cov_ff}
    \operatorname{cov}(f_{k,m}(t), f_{k',m'}(t')) = \sum_{i=1}^I \int_{-\infty}^{\infty} g_{k,m,i}(t - \tilde t; \btheta_{k,m,i}) \int_{-\infty}^{\infty} g_{k',m',i}(t' - \tilde t'; \btheta_{k',m',i})h_{u_i}(\tilde t, \tilde t')d\tilde t d\tilde t'.  
\end{equation} 
In practice, one can derive closed-form expressions for the covariances \eqref{eq:cov_fu} and \eqref{eq:cov_ff} by choosing the radial basis function (RBF) kernel for the latent GPs and the Gaussian smoothing kernel. For more details, refer to~\cite{chung2024federated}.

Despite that the CP-based approach can build a valid covariance for MGPs, it is apparent that sites $k$ and $k'$ still need to disclose their local data $\bt_{k,m}$ and $\bt_{k',m'}$ to calculate the cross-covariance $\operatorname{cov}(f_{k,m}(t), f_{k',m'}(t'))$, which is restrictive in federated scenarios. This issue is addressed by the inducing point approach. The inducing point approach was originally proposed to tackle the notorious inscalability of MGPs~\citep{alvarez2008sparse}. Interestingly, the sparse approximation with inducing points in turn enables avoiding the direct calculation of the cross-covariance across units at different sites. 

The inducing point approach starts with appreciating the conditional independence of $f_{1,1},\ldots,f_{K,M}$ given the shared latent functions $u_1,\ldots,u_I$ over the entire domain of $t$. This property is a direct consequence of the CP-based MGP construction, where functions $f_{1,1},\ldots,f_{K,M}$ depend only on unit-specific kernel $h_{k,m,i}$ if we know $u_1,\ldots,u_I$ perfectly. The inducing point approach assumes that the conditional independence would still hold even if we are given knowledge on $u_1,\ldots,u_I$ at a set of inducing points $\bz_i = [z_{i,p}]^T_{p=1,\ldots,P_i} \in \bbR^{P_i\times 1}$ for $i=1,\ldots,I$ on the entire domain. 

To formulate, consider the vector of function values $\bff_{k,m} = [f_{k,m}(t_{k,m,l})]^T_{l=1,\ldots,L_{k,m}} $ and $\bu_i = [u_i(z_{i,p})]^T_{p=1,\ldots,P_i}$; collectively, $\bff = [\bff^T_{k,m}]^T_{k=1,\ldots,K,\; m=1,\ldots,M_k}\in \bbR^{L \times 1} $, $\bu = [\bu^T_i]^T_{i=1,\ldots,I} \in \bbR^{P \times 1}$, and $\bz = [\bz^T_i]^T_{i=1,\ldots,I}\in \bbR^{P \times 1}$ with $P=\sum_{i=1}^I P_i$. Modeling $\bu_1,\ldots,\bu_I$ as independent GPs, the conditional independence of $\bff_{1,1},\ldots,\bff_{K,M}$ given $\bu_1,\ldots,\bu_I$ is represented as 
\begin{align}
   p(\bff\vert \bu, \bt, \bz) &= \prod_{k=1}^{K} \prod_{m=1}^{M_k} p(\bff_{k,m} \vert \bu, \bt_{k,m}, \bz) = \prod_{k=1}^{K} \prod_{m=1}^{M_k}\cN(\bff_{k,m}; \bK_{\bff_{k,m}, \bu}\bK_{\bu,\bu}^{-1}\bu, \bK_{\bff_{k,m}, \bff_{k,m}} -  \bK_{\bff_{k,m}, \bu} \bK_{\bu,\bu}^{-1}\bK_{\bu, \bff_{k,m}}), \label{eq:p(f|u)}\\
   p(\bu\vert \bz) &= \cN(\bu; \zeros, \bK_{\bu,\bu}) = \prod_{i=1}^I \cN(\bu_i;\zeros, \bK_{\bu_i, \bu_i}), \label{eq:p(u|z)}
\end{align}
where $\bK_{\bff_{k,m}, \bu} = \bK_{\bu, \bff_{k,m}}^T = [\bK_{\bff_{k,m}, \bu_i}]_{i=1}^I$ and $\bK_{\bu, \bu} = \operatorname{bdiag}(\bK_{\bu_i, \bu_i})_{i=1}^I$ with ``bdiag'' being a block diagonal matrix. 
The covariance matrices $\bK_{\bff_{k,m}, \bff_{k,m}}$, $\bK_{\bff_{k,m}, \bu_i}$, and $\bK_{\bu_i, \bu_i}$ are characterized by covariances $\operatorname{cov}(f_{k,m}(t), f_{k,m}(t'))$ in \eqref{eq:cov_ff}, $\operatorname{cov}(f_{k,m}(t), u_i(z))$ in \eqref{eq:cov_fu}, and $ \operatorname{cov}(u_i(z), u_i(z'))$ with the kernel $h_{u_i} (z, z'; \bxi_i)$, respectively. Here the notation $\bK_{(\cdot, \cdot)}$ indicates a (cross) covariance matrix between two random variables appearing at the subscript. For instance, $\bK_{\bbf_{k,m}, \bbf_{k',m'}} \in \bbR^{V_{k,m} \times V_{k',m'}}$ denotes the cross covariance matrix for $\bbf_{k,m}$ and $\bbf_{k',m'}$ (i.e., simply the covariance of $\bbf_{k,m}$ if $m = m'$ and $k=k'$).

Given \eqref{eq:func}, \eqref{eq:p(f|u)}, and \eqref{eq:p(u|z)}, we can derive a marginal likelihood, written as 
\begin{align}\label{eq:p(y|x,z)}
    p(\by \vert \bt, \bz) = \int p(\by \vert \bff)  p(\bff \vert \bu, \bt, \bz) p(\bu \vert \bz)d\bff d\bu = \cN(\by; \zeros, \bOmega +  \bK_{\bff, \bu} \bK_{\bu,\bu}^{-1}\bK_{\bu, \bff}+\bSigma),
\end{align}
where $\bOmega = \operatorname{bdiag}(\bK_{\bff_{k,m}, \bff_{k,m}} -  \bK_{\bff_{k,m}, \bu} \bK_{\bu,\bu}^{-1}\bK_{\bu, \bff_{k,m}})_{k=1, \; m=1}^{K, \; M_k}$, $\bK_{\bff, \bu} = \bK_{\bu, \bff}^T = [\bK^T_{\bff_{k,m}, \bu}]^T_{k=1,\ldots,K, \; m=1,\ldots,M_k}$, and $\bSigma$ indicates a diagonal matrix for noise variance, where its diagonal vector is $[\sigma^2_{k,m} \ones^T_{L_{k,m}}]^T_{k=1,\ldots,K, \; m=1,\ldots,M_k} \in \bbR_+^{M}$.

Here, it is crucial to note that the MGP in $\eqref{eq:p(y|x,z)}$ does not contain any cross-covariance $\bK_{\bff_{k,m},\bff_{k',m'}}$ with $k\neq k'$ for which calculation necessitates sharing $\bt_{k,m}$ and $\bt_{k',m'}$ between sites $k$ and $k'$. Instead, it has an approximated covariance $\bOmega + \bK_{\bff, \bu} \bK_{\bu,\bu}^{-1}\bK_{\bu, \bff}$ built upon pseudo inputs $\bz$. Therefore, sites $k$ and $k'$ share pseudo inputs $\bz$ with each other in place of $\bt_{k,m}$ and $\bt_{k',m'}$, thus allowed to secure their local data.

\paragraph{Extrapolation for time-varying observations} Given the MGP above, we now can extrapolate the time-dependent observations of unit $m$ at site $k$ by deriving a predictive distribution of function $f_{k,m}(\cdot)$ at an any arbitrary input $t$. In the spirit of Bayesian modeling, deriving a predictive distribution involves a posterior for the latent variables $\bff$ and $\bu$. Specifically, we consider a posterior distribution $p(\bff,\bu| \by,\bt, \bz) = p(\bff| \bu, \bt, \bz) p(\bu|\by, \bt,\bz)$. If we assume that the posterior $p(\bu|\by, \bt,\bz)$ can be inferred as a Gaussian $\cN(\bu; \hat \bmu_{\bu}, \hat \Psi_{\bu,\bu})$ with an estimated mean $\hat \bmu_{\bu}$ and covariance $\hat \bPsi_{\bu,\bu}$, the predictive distribution of function $f_{k,m,*} := f_{k,m}(t_{*})$ at a new arbitrary input $t_{*}$ is derived as 
\begin{align}\label{eq:pred}
    &p\left(f_{k,m,*} \vert t_{*}, \by, \bt, \bz\right) \nonumber\\
    &\qquad= \int p\left(f_{k,m,*} \vert \bu, t_*, \bt, \bz \right)p\left(\bu\vert\by, \bt, \bz\right)d\bu \nonumber\\
    &\qquad= \cN(f_{k,m,*};\bK_{f_{k,m,*}, \bu}\bK_{\bu,\bu}^{-1}\hat \bmu_{\bu}, \operatorname{cov}(f_{k,m}(t_*), f_{k,m}(t_*)) +  \bK_{f_{k,m,*}, \bu} (\hat \bPsi_{\bu,\bu} -\bK_{\bu,\bu}^{-1})\bK_{\bu, f_{k,m,*}}),
\end{align}
where $\operatorname{cov}(f_{k,m}(t_*), f_{k,m}(t_*))$ and $\bK_{f_{k,m,*}, \bu}$ are characterized by \eqref{eq:cov_ff} and \eqref{eq:cov_fu}, respectively.

One can see that, to derive \eqref{eq:pred}, we need to infer the posterior distribution $p(\bu|\by, \bt,\bz)$ as well as all hyperparameters $\bTheta_\text{hyp} :=  \{\{\btheta_{k,m,i}\}_{k=1, m=1, i=1}^{K, M_k, I}, \{\sigma_{k,m}\}_{k=1,m=1}^{K,M_k}, \{\bxi_i\}_{i=1}^I \}$ of the kernels involved in~\eqref{eq:p(y|x,z)}. This step can be viewed as a training process of the MGP. Unfortunately, it is not trivial in a federated scenario. The challenge is attributed to the inference of the posterior distribution $p(\bu|\by, \bt, \bz)$, which depends on all time-varying data $\left\{\bt_{k,m}, \by_{k,m}\right\}_{k=1,m=1}^{K,M_k}$ distributed across sites where data sharing is restrictive. Furthermore, estimating $\bTheta_\text{hyp}$ using the maximum likelihood estimation method requires maximizing the logarithm of \eqref{eq:p(y|x,z)}, which is not feasible with well-known FL approaches that assume the objective function can be separable across sites, for example, as a linear combination of independent terms for each site $k=1,\ldots,K$. We will discuss how these challenges can be addressed in Section~\ref{sec:inference}.

\subsubsection{Modeling time-to-failure events using the CoxPH model}
\label{sec:fed_survival}

Once we fit the FedMGP model $\hat{f}_{k,m}(t)$ to the observed CM signals $\by_{k,m}$ for all sites $k = 1,\ldots,K$ and its units $m = 1,\ldots,M_k$, as described in Section~\ref{sec:fed_mgp}, we further analyze the time-to-failure data $(V_{k,m},\delta_{k,m})$ and time-invariant covariates $\bm{w}_{k,m}$ by relating them to each unit's signal through the hazard function $h_{k,m}(t)$ as 
\begin{equation}
\label{eq:coxph}
    h_{k,m}(t) = h_0(t) \exp \left\{ \bm{\gamma}^T\bm{w}_{k,m} + \beta \hat{f}_{k,m}(t) \right\}, \; 
\end{equation}
where $h_0(t)$ is a baseline hazard function, shared by all units, with parameters $\bm{\lambda}$. This baseline function is typically modeled by the exponential,  Weibull, or a piecewise constant function, depending on the problem under study. Here, $\bm{\gamma}$ is a vector of coefficients for $\bm{w}_{k,m}$, and $\beta$ is a scaling parameter for $\hat{f}_{k,m}(t)$. A set of parameters, $ \bm{\phi}= \{\bm{\lambda}, \bm{\gamma}, \beta\}$, is to be estimated by maximizing the full log-likelihood function with a scaling factor $1/M$, expressed as 
\begin{equation}
\begin{aligned}
    \label{eq:log_likelihood}
    \ell(\bm{\phi}) &= \dfrac{1}{M} \sum_{k=1}^K\sum_{m=1}^{M_k} \log p(V_{k,m},\delta_{k,m} |\bm{w}_{k,m}, \hat{f}_{k,m}; \bm{\phi}) \\ &=  
    \dfrac{1}{M} \sum_{k=1}^K\sum_{m=1}^{M_k} \Bigg[   
    \delta_{k,m} \log 
    \left[h_0 (V_{k,m}) \exp \left\{ \bm{\gamma}^T \bm{w}_{k,m} + \beta\hat{f}_{k,m}(V_{k,m}) \right\}\right] - \int_{0}^{V_{k,m}} h_0(u) \exp \left\{ \bm{\gamma}^T \bm{w}_{k,m} + \beta\hat{f}_{k,m}(V_{k,m}) \right\} du \Bigg],
\end{aligned}
\end{equation}
where $ p(V_{k,m},\delta_{k,m} |\bm{w}_{k,m}, \hat{f}_{k,m}; \bm{\phi})$ for $m$th unit at site $k$ is  
\begin{equation}
\begin{aligned}
    \label{eq:likelihood}
    p(V_{k,m},\delta_{k,m} |\bm{w}_{k,m}, \hat{f}_{k,m}; \bm{\phi}) &= \left[h_{k,m}(V_{k,m}|\bm{w}_{k,m}, \hat{f}_{k,m}; \bm{\phi})\right]^{\delta_{k,m}} S (V_{k,m}|\bm{w}_{k,m}, \hat{f}_{k,m}; \bm{\phi}),
\end{aligned}
\end{equation}
with $S(t)=\mathbb{P}(T \ge t)$ being the survival function that represents the probability of survival up to time $t$. Since the hazard function $h(t)$ can be expressed by $h(t)=\lim_{\Delta t \rightarrow 0} \mathbb{P} (t < T \le t + \Delta t | T \ge t)/\Delta t = - \frac{d}{dt} \log S(t)$ as the instantaneous rate of occurrence of an failure event at time $t$, given that the event has not occurred before time $t$, we can relate $S(t)$ to $h(t)$ as the pdf $p(t) = - S'(t) = h(t) S(t)$, so that $S(t) = \mathbb{P}(T \ge t) = \exp \left\{-\int_{0}^t h(u) du \right\}$. If the event is observed (i.e., $\delta_{k,m}=1$), the contribution to the likelihood is $p(t)$, whereas if the event is censored (i.e., $\delta_{k,m}=0$), its contribution can be expressed as $S(t)$. Thus, the full likelihood can be constructed using all training units with respect to $\bm{\phi}$. Overall, the formula in~\eqref{eq:likelihood} captures the likelihood contribution in survival analysis, reflecting two possible outcomes we considered in this paper: failure or censorship. If the failure event occurs, the likelihood indicates surviving until time $t$ and failing at time $t$. If the unit is censored, the likelihood reduces to $S (V_{k,m}|\bm{w}_{k,m}, \hat{f}_{k,m}; \bm{\phi})$, as survival until time $t$ is the only observed information. More details about the CoxPH model and its likelihood derivation can be found in~\cite{kalbfleisch2011statistical}.

Note that calculating the integral in~\eqref{eq:log_likelihood} is computationally intensive, and it lacks a closed-form solution. Therefore, it can be approximated using numerical methods such as Gauss-Hermite quadrature, Gauss-Legendre quadrature, or other standard techniques like Simpson's method~\citep{zhou2014remaining}.

It should also be noted that we use the full likelihood rather than the partial likelihood for parameter estimation for $\bphi$. This is because the partial likelihood function is not clearly applicable in a FL setting, where the objective function needs to be separable by site. We will show how to apply the FL framework to train the survival model  momentarily in Section~\ref{sec:inference}.

\paragraph{Estimation of failure probabilities} After estimating the parameters, $\hat{\bm{\phi}} = \{\hat{\bm{\lambda}}, \hat{\bm{\gamma}}, \hat{\beta}\}$, by federated inference in Section~\ref{sec:inference}, we can estimate the conditional failure probability, $\hat{F}_{k,m}(t^*+\Delta t | t^*) := \hat{F}(t^*+\Delta t | t^*, \bm{w}_{k,m}, \hat{f}_{k,m}; \hat{\bm{\phi}})$, at time $t^*$ for $\Delta t$-ahead prediction for each unit $m$ at site $k$, provided that the observations of each unit have been recorded up to time $t^*$. Thus, we can calculate $\hat{F}_{k,m}(t^*+\Delta t | t^*)$ as one minus estimated conditional survival probability, $1-\hat{S}(t^*+\Delta t | t^*, \bm{w}_{k,m}, \hat{f}_{k,m}; \hat{\bm{\phi}})$, as follows.

\begin{equation}
\begin{aligned}
    \label{eq:failure_prob_predict}
    \hat{F}_{k,m}(t^*+\Delta t | t^*) &:=
    \hat{F}(t^*+\Delta t | t^*, \bm{w}_{k,m}, \hat{f}_{k,m}; \hat{\bm{\phi}}) \\ &= 1 - \hat{S}(t^*+\Delta t | t^*, \bm{w}_{k,m}, \hat{f}_{k,m}; \hat{\bm{\phi}}) \\
    &= 1- \dfrac{\hat{S}(t^*+\Delta t |\bm{w}_{k,m}, \hat{f}_{k,m}; \hat{\bm{\phi}})}{\hat{S}(t^*|\bm{w}_{k,m}, \hat{f}_{k,m}; \hat{\bm{\phi}})} \\
    &= 1- \exp \left[- \int_{t_*}^{t^* + \Delta t} \hat{h}_0 (u) \exp \left\{ \hat{\bm{\gamma}}^T \bm{w}_{k,m} + \hat{\beta} \hat{f}_{k,m}(u) \right\} du \right],
\end{aligned}
\end{equation}
where $\hat{S}(\cdot)$ is the estimated survival probability.

\subsection{Federated Inference for Fed-Joint}
\label{sec:inference}

We discuss how to infer the CoxPH model in the federated scenario and predict the future failure time and probabilities using this estimated survival model. Suppose the raw data cannot be shared across the sites, but only the respective model or its parameters trained at each site can be shared through the central repository due to privacy concerns. Suppose further that the communication costs are relatively high for transferring data from local sites to the central repository.


\paragraph{Federated Inference for MGP} We first show how to estimate the MGP in a federated manner. Our goal is to estimate the posterior distribution $p(\bu \vert \by, \bt, \bz)$ and the hyperparameters $\bTheta_{\sf hyp}$, while the data $\{\bt, \by\}$ are distributed across different sites, with restrictions on data sharing. The key idea is to formulate an optimization problem for estimating the posterior and hyperparameters in the FL setting. Specifically, we resort to stochastic variational inference (VI)~\citep{hoffman2013stochastic}. As a special case of VI, stochastic VI introduces a variational distribution, which is often chosen to have a manageable form (e.g., factorized Gaussians), to approximate the exact posterior. Estimating the variational distribution involves minimizing its Kullback-Leibler (KL) divergence from the posterior. It is known that minimizing the KL divergence is equivalent to maximizing a lower bound of the marginal log-likelihood, known as the evidence lower bound (ELBO). Thus, by solving an optimization problem that maximizes the ELBO with respect to the variational distributions and the hyperparameters of MGPs, we can obtain an approximate posterior and estimate the MGP hyperparameters. In particular, stochastic VI enables us to construct an objective function that is separable across each observation, making it compatible with FL.

Consider a variational distribution $q(\bff, \bu) = p(\bff | \bu, \bt, \bz) q(\bu) = \prod_{k=1}^{K}\prod_{m=1}^{M_k}p(\bff_{k,m} | \bu, \bt, \bz)q(\bu)$ to approximate $p(\bff, \bu \vert \by, \bt, \bz)$, where we let $q(\bu) := \cN(\bu ; \bmu_\bu, \bPsi_{\bu, \bu}) = \prod_{i=1}^I q(\bu_i) = \prod_{i=1}^I \cN(\bu_i ; \bmu_{\bu_i}, \bPsi_{\bu_i, \bu_i})$ to surrogate $p(\bu \vert \by, \bt, \bz)$. Given this approximation, we can derive the surrogate of $p(\bff_{k,m} \vert \by, \bt, \bz)$ as
\begin{align}
    q(\bff_{k,m} \vert \bt, \bz) = \int p(\bff_{k,m} | \bu, \bt, \bz)q(\bu) d\bu = \cN(\bff_{k,m} ; \bK_{\bff_{k,m}, \bu}\bK_{\bu,\bu}^{-1}\bmu_{\bu},  \bK_{\bff_{k,m}, \bff_{k,m}} +  \bK_{\bff_{k,m}, \bu} (\bPsi_{\bu,\bu} -\bK_{\bu,\bu}^{-1})\bK_{\bu, \bff_{k,m}}).
\end{align}
With the variational distributions, the ELBO is derived as follows.
\begin{align}
    \log p(\by \vert \bt, \bz) = \log \int p(\by \vert \bff) p(\bff, \bu \vert \bt, \bz) d\bff d\bu \ge \int q(\bff, \bu) \log \frac{p(\by \vert \bff) p(\bff, \bu \vert \bt, \bz)}{q(\bff, \bu)}  d\bff d\bu = \cL_{\sf ELBO} (\bTheta),
\end{align}
where the inequality is due to Jensen's inequality. To estimate the model, $\cL_{\sf ELBO} (\bTheta)$ is maximized with respect to the parameter $\bTheta := \{\bTheta_{\sf hyp}, \bTheta_{\sf var}\}$ where $\bTheta_{\sf var} = \{\bmu_{\bu}, \bPsi_{\bu, \bu} \}$ associated with the variational distributions introduced. The ELBO can be further derived as 
\begin{align}\label{eq:ELBO}
\cL_{\sf ELBO}(\bTheta) = \sum_{k=1}^{K}\sum_{m=1}^{M_k}\bbE_{q(\bff_{m,k})} \left[ \log p (\by_{m,k} \vert \bff_{m,k}) \right] - \sum_{i=1}^I \KL (q(\bu_i)\Vert p(\bu_i)) = \sum_{k=1}^K \cV_k(\bTheta_k, \bTheta^{\sf glob}),
\end{align}
where
\begin{align}
\cV_k(\bTheta_k, \bTheta^{\sf glob}) = \sum_{m=1}^{M_k}\bbE_{q(\bff_{m,k})} \left[ \log p (\by_{m,k} \vert \bff_{m,k}) \right] - r_k\sum_i^I \KL(q(\bu_i)\Vert p(\bu_i)),
\end{align}
with $r_k = \sum_{m=1}^{M_k}L_{k,m}/(\sum_{k=1}^{K}\sum_{m=1}^{M_k}L_{k,m})$  proportional to the number of observations at each site $k$ and reorganizing the parameters $\bTheta = \{\{\bTheta_k\}_{k=1}^K, \bTheta^{\sf glob}\}$ to denote site-specific parameters $\bTheta_k := \{\{\btheta_{k,m,i}\}_{m=1;i=1}^{M_k;I}, \{\sigma_{k,m}\}_{m=1}^{M_k} \}$ and global parameters $\bTheta^{\sf glob} = \{ \{\bxi_i\}_{i=1}^I, \bmu_{\bu}, \bPsi_{\bu, \bu} \}$. 

Importantly, \eqref{eq:ELBO} clearly shows the separability across sites. Each term $\cV_k(\cdot)$ is independent of the data of any other sites $k' \neq k$. The separability allows us to employ an FL algorithm to maximize $\cL_{\sf ELBO}(\bTheta)$. More precisely, each site $k$ takes local update steps to minimize  $\tilde{\cV}_k(\bTheta_k, \bTheta^{\sf glob}) := - \cV_k(\bTheta_k, \bTheta^{\sf glob})$ with respect to $\bTheta_k$ and $\bTheta^{\sf glob}$. Then the sites share locally updated $\bTheta_k^{\sf glob}$, for all $k=1,\ldots,K$, through the central server, and the server aggregates them to obtain $\bTheta^{\sf glob}$ as in~\eqref{eq:parameter_aggregate}. The aggregated parameter $\bTheta^{\sf glob}$ is then distributed to each site to update $\bTheta_k^{\sf glob}$ locally, along with $\bTheta_k$. This process is iterated until a termination condition is met. This is represented in lines~{3-11} of Algorithm \ref{al:fed_global}.

\paragraph{Federated Inference for the CoxPH model} To estimate $\bm{\phi}$, we maximize the scaled log-likelihood function $\ell(\bm{\phi})$ in~\eqref{eq:log_likelihood}. For convenience, we minimize the negative (scaled) log-likelihood function $\tilde{\ell}(\bm{\phi}) := -\ell(\bm{\phi})$. Since $\tilde{\ell}(\bm{\phi})$ is separable in terms of sites and units, we can express $\tilde{\ell}(\bm{\phi})$ as the finite average of the objectives $\tilde{\ell}_{k,m}(\bm{\phi})$ of the form 
\begin{equation}
\begin{aligned}
\label{eq:objective}
    \tilde{\ell}(\bm{\phi}) = \dfrac{1}{M} \sum_{k=1}^{K} \sum_{m=1}^{M_k} \tilde{\ell}_{k,m}(\bm{\phi})
    =  \sum_{k=1}^{K} \dfrac{M_k}{M} \tilde{\ell}_{k}(\bm{\phi}),
\end{aligned}
\end{equation}
where $\tilde{\ell}_{k}(\bm{\phi}) = \sum_{m = 1}^{M_k} \tilde{\ell}_{k,m}(\bm{\phi}) / M_k$ with $\tilde{\ell}_{k,m}(\bm{\phi}) = \log p(V_{k,m},\delta_{k,m} |\bm{w}_{k,m}, \hat{f}_{k,m}; \bm{\phi})$ for all $k=1,\ldots,K$.

Then each site $k$ takes local update steps to minimize  $\tilde{\ell}_{k}(\bm{\phi})$ with respect to $\bm{\phi}$, so we have the locally updated parameters $\bm{\phi}_k$ for all $k$. Thus, $\tilde{\ell}_{k}(\bm{\phi})$ is separable across sites. The sites upload the locally updated $\bm{\phi}_k$ to the central server, and the server aggregates them to obtain $\bm{\phi}$ as in~\eqref{eq:parameter_aggregate}. The aggregated parameter $\bm{\phi}$ is then distributed to each site to update $\bm{\phi}_k$ locally. This process is iterated until a termination condition is met. This is represented in lines~{13-21} of Algorithm \ref{al:fed_global}.

We summarize the Fed-Joint algorithm, which operates the two-stage estimation procedure, via Algorithm~\ref{al:fed_global}, by combining the two federated inference procedures described above in order to sequentially infer the parameters $\bm{\Theta}$ and $\bm{\phi}$ for the MGP and CoxPH models.

\begin{algorithm}[htb!] 
\caption{Federated Joint Modeling of Degradation Signals and Failure Events (Fed-Joint)} 
\label{al:fed_global}
	\begin{algorithmic}[1]
	    \STATE \textbf{Input:} initial parameters for $\bm{\Theta}=\{\{\bTheta_k\}_{k=1}^K,\bTheta^{\sf glob}\}$ and $\bm{\phi}$, step size $\eta_1$ and $\eta_2$, local iteration number $E_1$ and $E_2$, total communication rounds $R_1$ and $R_2$, prediction time $t^*$, and prediction horizon $\Delta t$.
        \STATE /* \textit{Federated Inference for MGP} */
        \FOR{round $r_1=1,\ldots,R_1$}
        \STATE The central server broadcasts $\bTheta^{\sf glob}$ to all sites $k=1,\ldots,K$ and set $\bTheta_k^{\sf glob} \leftarrow \bTheta^{\sf glob}$.
        \FOR{each site $k=1,\ldots,K$}
        \STATE Update the local model via
            \{$\bTheta_k,\bTheta^{\sf glob}_k\} \leftarrow $ \texttt{local\_update} $(\bTheta_k,\bTheta_k^{\sf glob}; \mathcal{D}_{k},E_1)$.  
        \STATE Upload $\bTheta^{\sf glob}_k$ to the central server.
        \ENDFOR
        \STATE Update $\bTheta^{\sf glob}$ via $\bTheta^{\sf glob} \leftarrow $ \texttt{central\_update} $(\{ \bTheta^{\sf glob}_k\}_{k=1}^K)$. 
        \STATE Obtain $\hat{f}_{k,m}(t)$ for all $m$ at all sites $k$. 
        \ENDFOR
        \STATE /* \textit{Federated Inference for the CoxPH model} */
        \FOR{round $r_2=1,\ldots,R_2$}
        \STATE The central server broadcasts $\bm{\phi}$ to all sites $k=1,\ldots,K$ and set $\bm{\phi}_k \leftarrow \bm{\phi}$.
        \FOR{each site $k=1,\ldots,K$}
        \STATE Update the local model via 
            $\bm{\phi}_k \leftarrow $ \texttt{local\_update} $(\bm{\phi}_k; \mathcal{D}_{k},E_2)$.  
        \STATE Upload $\bm{\phi}_k$ to the central server.
        \ENDFOR
        \STATE Update $\bm{\phi} \leftarrow $ \texttt{central\_update} $(\{\bm{\phi}_k\}_{k=1}^K)$.
        \ENDFOR
        \STATE \textbf{Output:} estimated mean RUL, $ \widehat{mrl}_{test,m}(t^*)$, based on $\hat{S}(t|t^*)$ at time $t^*$ using~\eqref{eq:estimate_mrl}; estimated conditional failure probability, $\hat{F}_{k,m}(t^*+\Delta t|t^*)$, for $\Delta t$-head prediction at time $t^*$ using~\eqref{eq:failure_prob_predict} for the target unit $m$ at site $k$.
    \end{algorithmic} 
\end{algorithm}

We provide detailed explanations of Algorithm~\ref{al:fed_global} in a unified manner for parameter updating in the MGP and CoxPH models.

\begin{figure}[h!]
    \centering
\includegraphics[width=0.95\linewidth]{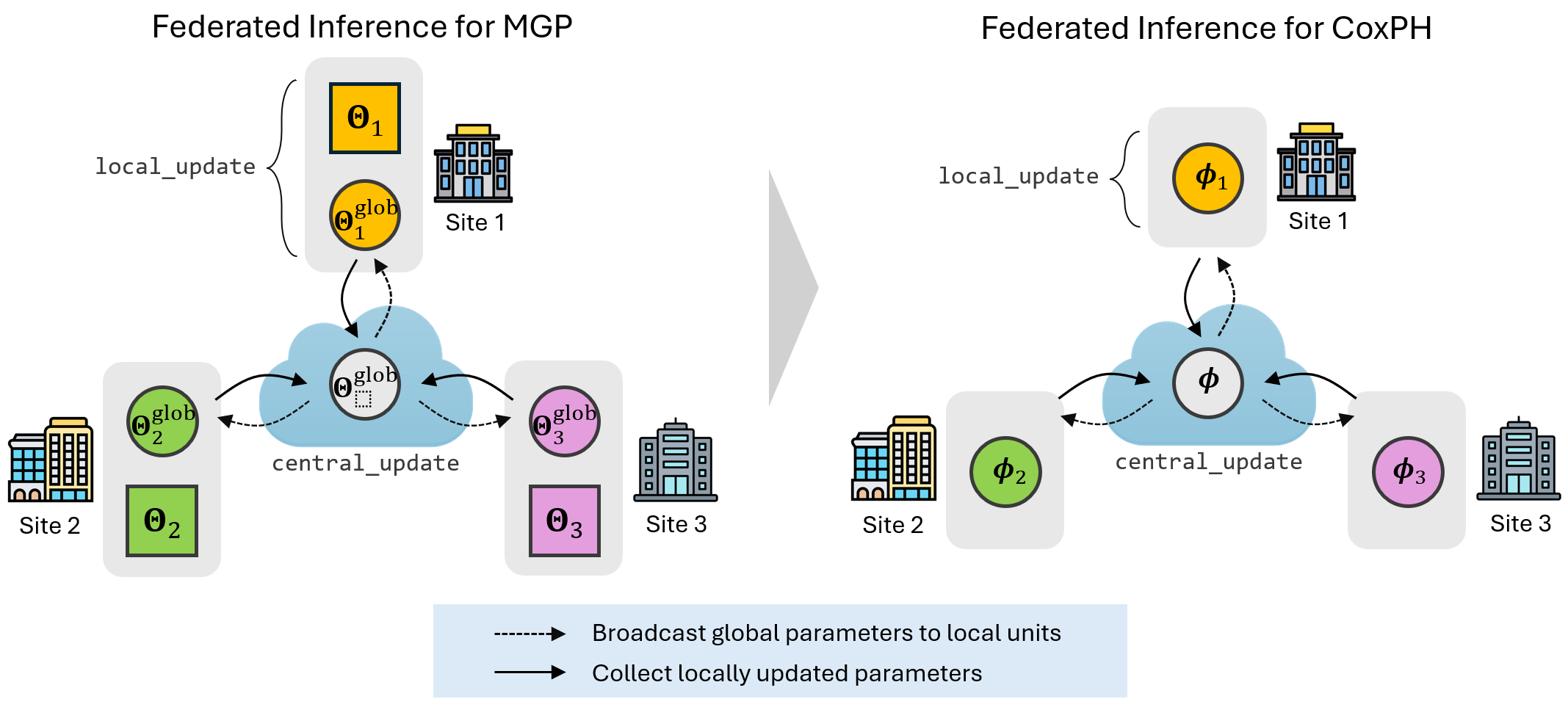}
    \caption{Parameter estimation procedure for Fed-Joint.}
    \label{fig:alg}
\end{figure}

\begin{itemize}
    \item Lines~5-8 and 15-18: Each site updates $\{\bTheta_k,\bTheta_k^{\sf glob}\}$ (resp., $\bm{\phi}_k$) via \texttt{local\_update}. 
    Specifically, \texttt{local\_update} minimizes the negative ELBO, $\tilde{\cV}_k(\bTheta_k,\bTheta_k^{\sf glob})$, with respect to $\{\bTheta_k,\bTheta_k^{\sf glob}\}$ (resp., negative log-likelihood function $\tilde{\ell}_k(\bm{\phi}_k)$ with respect to $\bm{\phi}_k$) using a gradient descent method that iterates 
    \begin{equation} \label{eq:parameter_update}
        \begin{bmatrix}
            \bTheta_k \\ \bTheta_k^{\sf glob} 
        \end{bmatrix} \leftarrow \begin{bmatrix}
            \bTheta_k \\ \bTheta_k^{\sf glob}
        \end{bmatrix} - \eta_1 \nabla \tilde{\cV}_k (\bTheta_k,\bTheta_k^{\sf glob}) \;\; (\text{resp., } \bm{\phi}_k \leftarrow \bm{\phi}_k - \eta_2 \nabla \tilde \ell (\bphi_k)).
    \end{equation}
    with $\eta_1, \eta_2 > 0$. After a sufficient number of iterations $E_1$ (resp., $E_2$), \texttt{local\_update} returns the updated global parameters $\bTheta^{\sf glob}_k$ (resp., $\bm{\phi}_k$) for each $k$. Note that the gradients can be derived analytically or estimated by finite-differencing. If the data size is relatively large in a specific site, one can randomly choose the subset of the data and conduct the (mini-batch) stochastic gradient descent method to update $\bTheta_k$,$\bTheta_k^{\sf glob}$, and $\bm{\phi}_k$.
    \item Lines~9 and 19: The central server receives the updated global parameters  $\bTheta^{\sf glob}_k$ (resp., $\bm{\phi}_k$) from each site and aggregates parameters to obtain $\bTheta^{\sf glob}$ (resp., $\bm{\phi}$), which is refer to as \texttt{central\_update}. For example, if \texttt{central\_update} uses a weighted average strategy, it returns the aggregated parameters as 
    \begin{equation} \label{eq:parameter_aggregate}
         \bTheta^{\sf glob} \leftarrow   \sum_{k=1}^K r_k\bTheta^{\sf glob}_k \;\; (\text{resp., } \bm{\phi}\leftarrow \sum_{k=1}^K \frac{M_k}{M}\bm{\phi}_k).
    \end{equation}   
\end{itemize}
The above procedure is repeated until some termination conditions are met or the prespecified communication rounds are attained. The two-stage estimation procedure outlined in Algorithm~\ref{al:fed_global} is illustrated in Figure~\ref{fig:alg}.

\section{Simulation Study}
\label{sec:simulation}

To demonstrate the effectiveness of the proposed Fed-Joint model, we conduct simulation studies using synthetic datasets.

\subsection{Data Generation}
\label{sec:simulation_data}

\begin{figure}[h]
\centering    
\subfigure[Scenario~I: polynomically growing signals]{\includegraphics[width=0.4\textwidth]{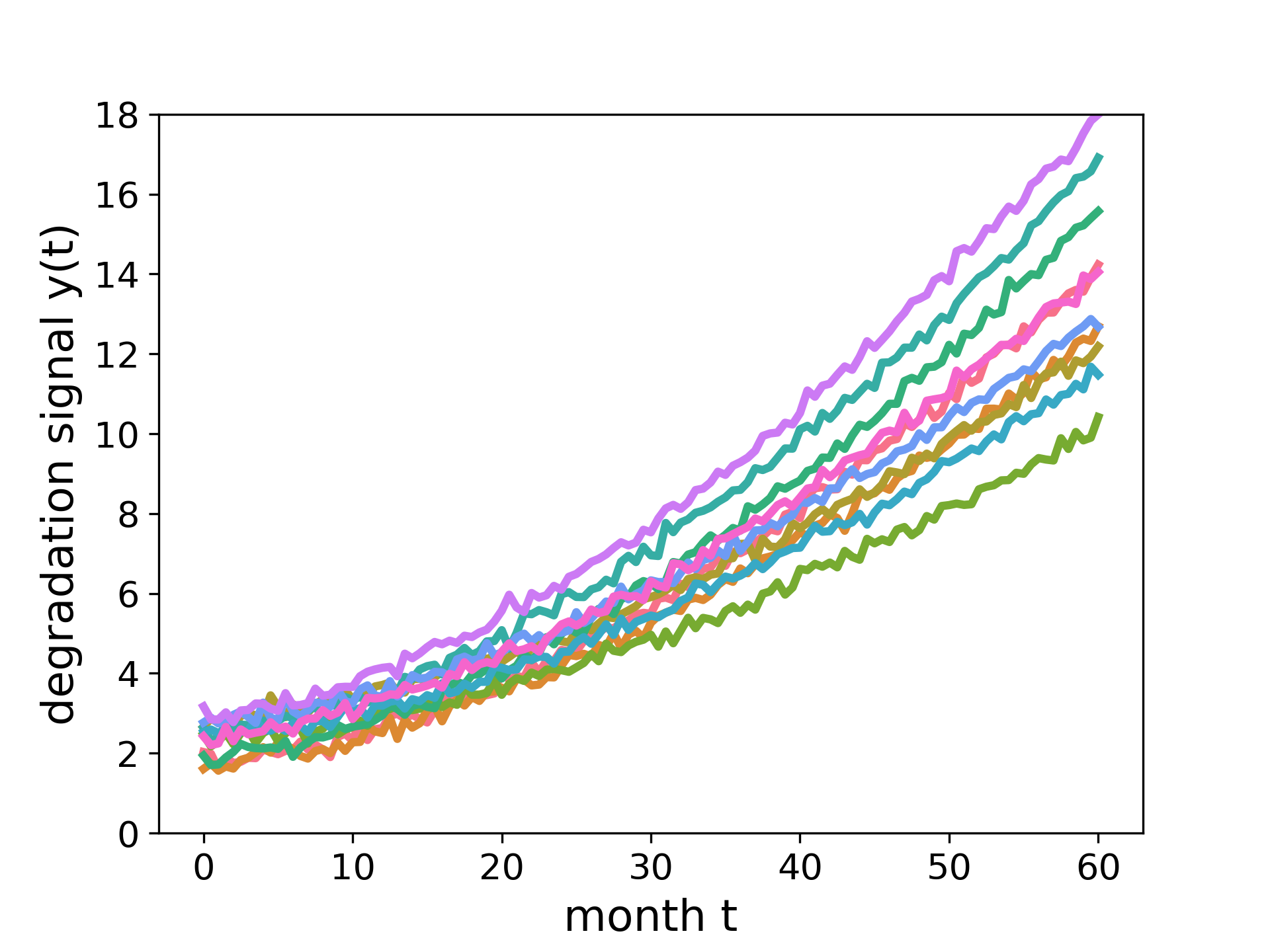}}
\subfigure[Scenario~II: highly nonlinear signals]{\includegraphics[width=0.4\textwidth]{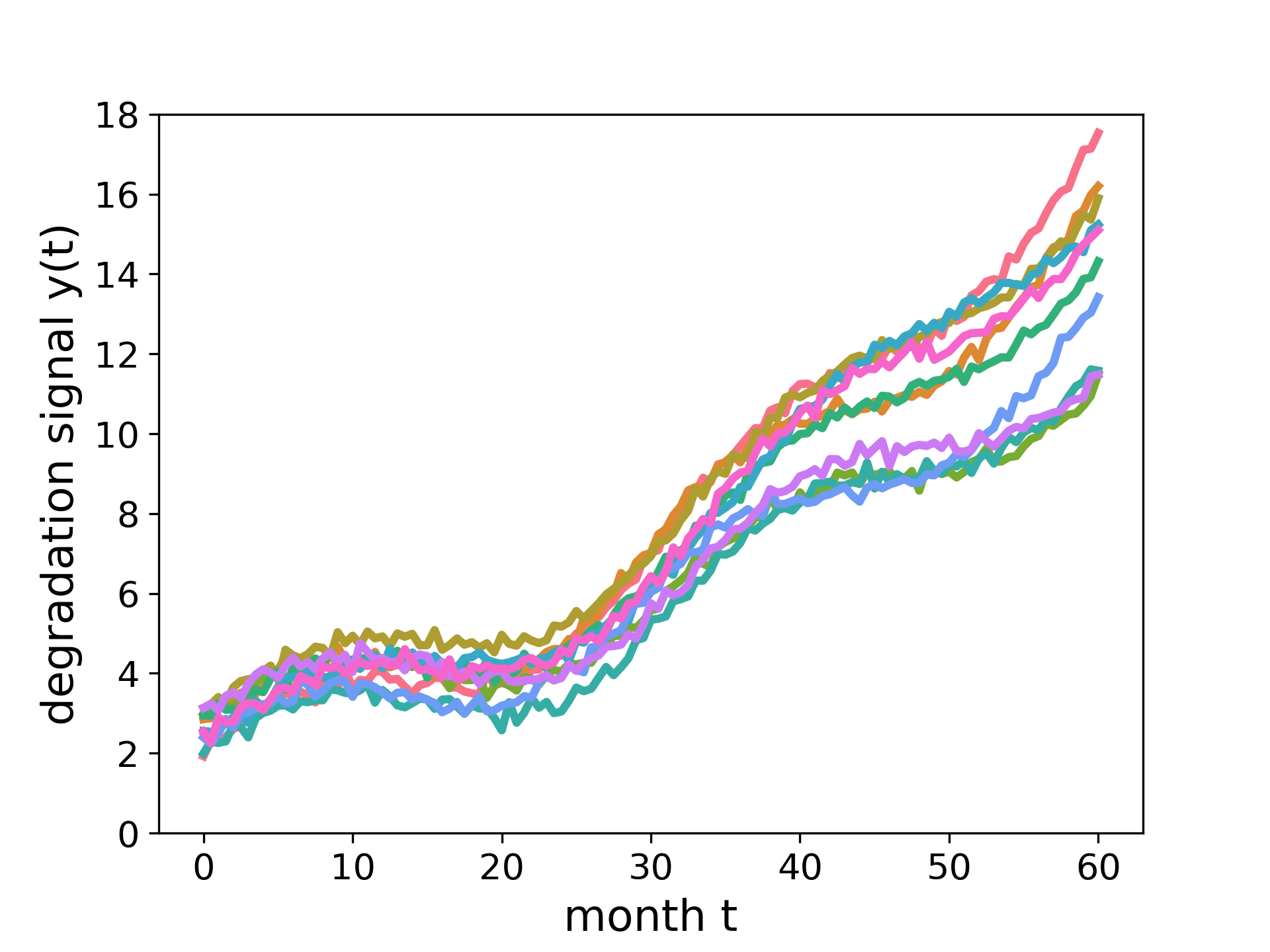}}
\caption{Examples of 10 randomly generated degradation signals for each scenario.}
\label{fig:degradation_signals}
\end{figure}

We consider two scenarios to evaluate whether Fed-Joint performs well, particularly in handling highly nonlinear degradation signals. For each scenario, we generate the data $\cD_{k,m} = \{V_{k,m}, \delta_{k,m}, \bt_{k,m}, \by_{k,m}, \bm{w}_{k,m}\}$ for all sites $k = 1,\ldots,K$ and units $m=1,\ldots,M_k$, with $(K, M_k)$ set to $(3, 20)$, $(3, 50)$, and $(5, 20)$. Without loss of generality, we assume that degradation signal values are observed at regularly spaced timestamps every 2 weeks, $\bt_{k,m} = [t_{k,m,1},\ldots, t_{k,m,L_{k,m}}]^T$, where $L_{k,m}=120$ for all sites and units. Specifically, we assume that the true degradation signals take the following form, with the white noise $\epsilon_{k,m}(t)$ superimposed.
\begin{equation}
\begin{aligned}
    \label{eq:true_signal}
    y_{k,m} (t) &= f_{k,m} (t) + \epsilon_{k,m}(t) \\
    &= b_{k,m,0} + b_{k,m,1} t^{1.2} + b_{k,m,2} t^{1.7} + z_{k,m}(t) + \epsilon_{k,m}(t),
\end{aligned}
\end{equation}
where $\epsilon_{k,m}(t) \sim N(0,0.2)$ and $z_{k,m}(t)$ plays a role in distinguishing between the two scenarios as 
\begin{equation}
    z_{k,m}(t) =
\begin{cases} 
    0 & \text{if Scenario~I}, \\
    c \cdot \sin(d t)  & \text{if Scenario~II},
\end{cases}
\end{equation}
with $c \sim \textit{unif}(0.99,1.01)$ and $d \sim \textit{unif}(0.18,0.22)$. 
Scenario~I represents a parametric underlying degradation signal that follows a similar setting as~\cite{zhou2014remaining}. Scenario II adds the sine term to make the signal wiggly, which allows us to test whether the proposed model works effectively for a highly nonlinear signal. This is especially important in cases where the form of the underlying signal is unknown \textit{a priori}, which is common in many real-world industrial applications. Also, a set of coefficients, $\bm{b}_{k,m} = [b_{k,m,0},b_{k,m,1},b_{k,m,2}]^T$, is simulated  through the multivariate normal $N(\bm{\mu}_b,\bm{\Sigma}_b)$ with mean $\bm{\mu}_b$ and covariance $\bm{\Sigma}_b$ as follows.
\begin{equation}
\begin{aligned}
    \bm{\mu}_b &= [2.5, 0.01, 0.01]^T, \\
    \bm{\Sigma}_b &= \begin{bmatrix}
0.2 & -4 \times 10^{-4} & 7 \times 10^{-5} \\
-4 \times 10^{-4} & 3 \times 10^{-6} & 10^{-7} \\
7\times 10^{-5} & 10^{-7} & 3 \times 10^{6}
\end{bmatrix}.
\end{aligned}
\end{equation}
Figure~\ref{fig:degradation_signals} shows the examples of randomly generated degradation signals described for Scenario~I and Scenario~II.

To employ the CoxPH model in~\eqref{eq:coxph}, we adopt the Weibull density function for a baseline hazard rate function $h_0(t)$, namely, $h_0(t) = \lambda \rho t^{\rho - 1}$ with the true parameters $\lambda = 0.001$ and $\rho=1.05$. We also assume that the time-invariant covariate $\bm{w}_{k,m}$ in~\eqref{eq:coxph} is a scalar dummy variable $w_{k,m}$ with $w_{k,m} = 1$ indicating that the unit is of Type~I, or $w_{k,m} = 0$ of Type~II for all $k$ and $m$, which are generated by the Bernoulli distribution with probability $0.5$, i.e., $w_{k,m} \sim \textit{Bernoulli}(0.5)$. Additionally, the true coefficients within the exponent in~\eqref{eq:coxph} are set as $\gamma = 0.2$ and $\beta = 0.5$. Thus, in the simulation studies, we utilize the following true hazard rate function.
\begin{equation}
    \label{eq:true_hazard}
    h_{k,m}(t) = h_0(t) \exp \left\{ 0.2 w_{k,m} + 0.5 f_{k,m}(t) \right\}.
\end{equation} 

For the event indicator $\delta_{k,m}$, we set $5\%$ of the units to be censored and assign $\delta_{k,m} = 0$; otherwise, $\delta_{k,m}=1$, i.e., $\delta_{k,m} \sim \textit{Bernoulli}(0.95)$. Moreover, to generate the failure time $T_{k,m}$, we draw a random sample that follows the true failure probability $F_{k,m}(t|0) \equiv F_{k,m}(t|0, w_{k,m}, f_{k,m}; \bm{\theta}) $  using inverse transform sampling: (i) generate a random sample $u$ from the standard uniform distribution on $[0,1]$, i.e., $U \sim \textit{unif}(0,1)$; (ii) approximate the inverse of $F_{k,m}(t|0)$, that is, $F_{k,m}^{-1}(t|0)$, using a linear interpolation method based on the inverse data $\{F_{k,m}(t|0),t\}_{t=0}^{t = t_{k,m,L_{k,m}}}$; (iii) compute the random variable $F_{k,m}^{-1}(u|0)$ and set $V_{k,m} = F_{k,m}^{-1}(u|0)$ as a failure time, with $\delta_{k,m} = 1$ for each $m$ and $k$; (iv) Among all units across all sites, set $5 \%$ of the units to be censored by setting $\delta_{k,m} = 0$ and set $V_{k,m} = t_{k,m,L_{k,m}}$; and (v) truncate $\bt_{k,m}$ at time $V_{k,m}$ and obtain the partial signal $\by_{k,m}$ up to the failure or censoring time $V_{k,m}$.  

Using the generated data $\cD_{k,m}$ for all $k = 1,\ldots,K$ and $m=1,\ldots,M_k$, with cases where $(K,M_k) = (3,20)$, $(3,50)$, and $(5,20)$ for two scenarios, we conduct the two-stage estimation method to obtain the MGP model $\hat{f}_{k,m}(t)$ by estimating $\{\bTheta,\bTheta^{\sf glob}\}$ and subsequently obtain the CoxPH model by estimating $\bm{\phi} = \{\bm{\lambda},\bm{\gamma},\beta\}$ in~\eqref{eq:coxph}, as described in Algorithm~\ref{al:fed_global}.

\subsection{Implementation}
\label{sec:simulation_implementation}

We compare the performance of Fed-Joint with three alternatives.

\begin{itemize}
    \item \textit{Cen-Joint}~\citep{yue2021joint}: This alternative assumes that all the data can be shared at the centralized site (or central server) without any communication cost. It uses the MGP~\citep{alvarez2008sparse} to fit the degradation signals and predicts failure times and probabilities based on this model in the joint modeling framework. It is expected to show the best performance, as it utilizes all the data to train the model, thereby providing the predictive competitiveness compared to Fed-Joint.
    \item \textit{Ind-Joint}: This benchmark assumes that the data cannot be shared across sites. Ind-Joint utilizes the same joint model as Cen-Joint, but it is implemented at each site individually.
    \item \textit{LMM-Joint}~\citep{zhou2014remaining}: This benchmark also assumes that the data cannot be shared across sites, so it is similar to Ind-Joint. However, compared to Ind-Joint, it adopts a linear mixed-effects model instead of MGP, assuming quadratic growth of the degradation signals, with the assumption that the signals follow a parametric form.
\end{itemize}

For each alternative along with Fed-Joint, we evaluate the predictive performance using the mean absolute error between the 
true RUL and estimated mean RUL for the testing units, denoted by $MAE_{mrl}$ expressed as  
\begin{equation}
    \label{eq:mae_mrl}
        MAE_{mrl} = \dfrac{1}{M_{test}} \sum_{m=1}^{M_{test}} \left| (T_{test,m} - t^*) - \widehat{mrl}_{test,m}(t^*) \right|,
\end{equation}
where $M_{test}$ is the number of units in the testing site, $T_{test,m} - t^*$ is the true RUL at time $t^*$, and $\widehat{mrl}_{test,m}(t^*)$ is the estimated mean RUL at time $t^*$, both for the unit $m$ in the testing site. $\widehat{mrl}_{test,m}(t^*)$ is calculated as 
\begin{equation}
    \label{eq:estimate_mrl}
    \widehat{mrl}_{test,m}(t^*) := \mathbb{E}[T-t^*|t^*] = \int_{t^*}^\infty \hat{S}_{test,m}(u|t^*) du.
\end{equation}

Another performance metric is the mean absolute error between the true and estimated failure probabilities, used to assess whether the failure probabilities obtained from each model are well estimated at time $t^*$ for $\Delta t$-ahead prediction for the units in the testing site. We call this metric $MAE_{F}$, expressed as  
\begin{equation}
    \label{eq:error_prob}
        MAE_{F} = \dfrac{1}{M_{test}} \sum_{m=1}^{M_{test}} \left| F_{test,m}(t^*+\Delta t|t^*) - \hat{F}_{test,m}(t^*+\Delta t|t^*) \right|,
\end{equation}
where $F_{test,m}(t^*+\Delta t|t^*)$ and $\hat{F}_{test,m}(t^*+\Delta t|t^*)$ are the true and estimated failure probabilities for $\Delta t$-ahead prediction at time $t^*$ for unit $m$, respectively, both of which are calculated using~\eqref{eq:failure_prob_predict}.

In our simulation studies, $M_{test} $ is $ 20$ or $50$, depending on the problem instance. Here, we set $t^*$ as the nearest upper time value of $\alpha \times V_{k,m}$, where $\alpha $ is the fraction between $0$ and $1$ that represents the percentage of observed data, taking values of $ 0.3, 0.5, 0.7$. The test set consists of the data from site $0$, and the remaining data from site $1$ to the final site is used as the training set. We conduct 20 repeated experiments with different training and test sets to evaluate the predictive performance of each method. 

\subsection{Results and Analysis}
\label{sec:simulation_result}

We summarize the prediction accuracy of each method in Tables~\ref{tb:mae_mrl}-\ref{tb:mae_prob2} in terms of the number of sites and units, the shape of the underlying degradation signals (i.e., Scenario I and Scenario II), and prediction time $t^*$ with fractions $\alpha=0.3, 0.5, 0.7$.

Before we compare the performance of the methods, we observe some common observations from Tables~\ref{tb:mae_mrl}-\ref{tb:mae_prob2}. First, the prediction accuracy tends to improve as $t^*$ increases (i.e., $\alpha$ increases) in all scenarios and methods, as seen in the diminishing values of both $MAE_{mrl}$ and $MAE_{F}$. This is because the accuracy of the fitted degradation signals improves with more observations. Second, as the number of units participating in model training increases, the prediction accuracy improves in all scenarios and methods, since more data are available for accurate parameter estimation in both degradation and survival models.

\begin{figure}[h]
\centering    
\subfigure[Fed-Joint]{\includegraphics[width=0.4\textwidth]{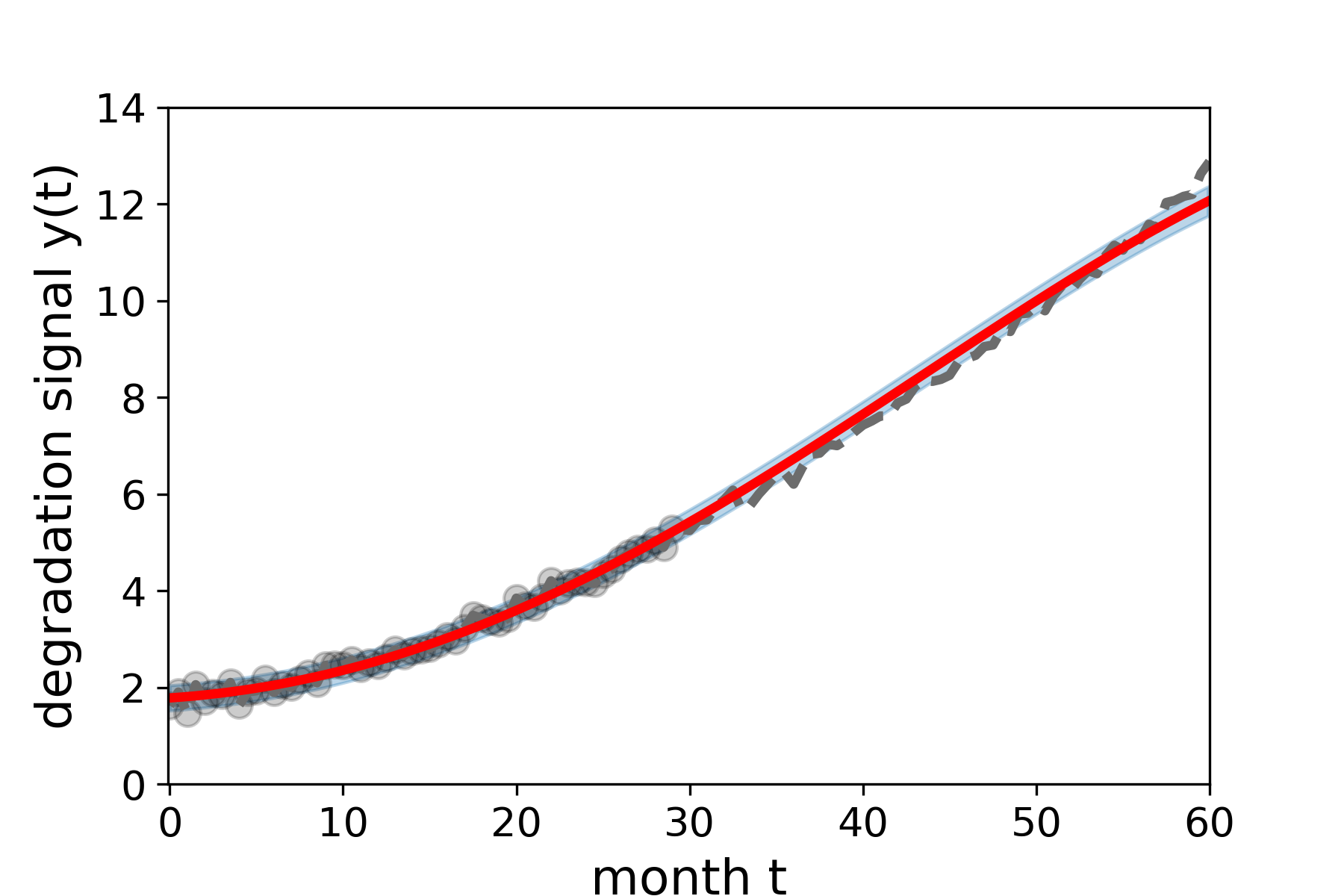}}
\subfigure[Cen-Joint]{\includegraphics[width=0.4\textwidth]{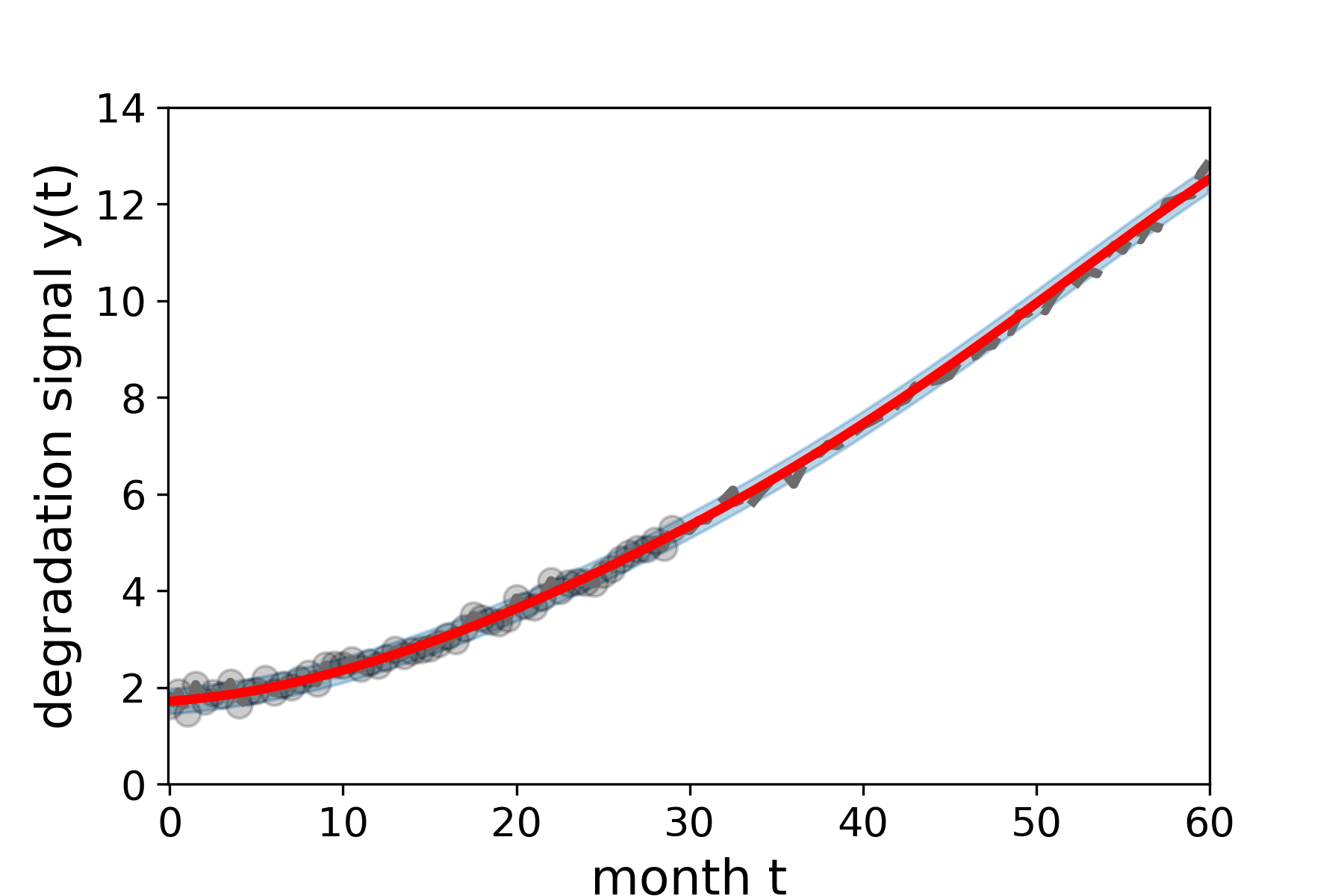}} 
\subfigure[Ind-Joint]{\includegraphics[width=0.4\textwidth]{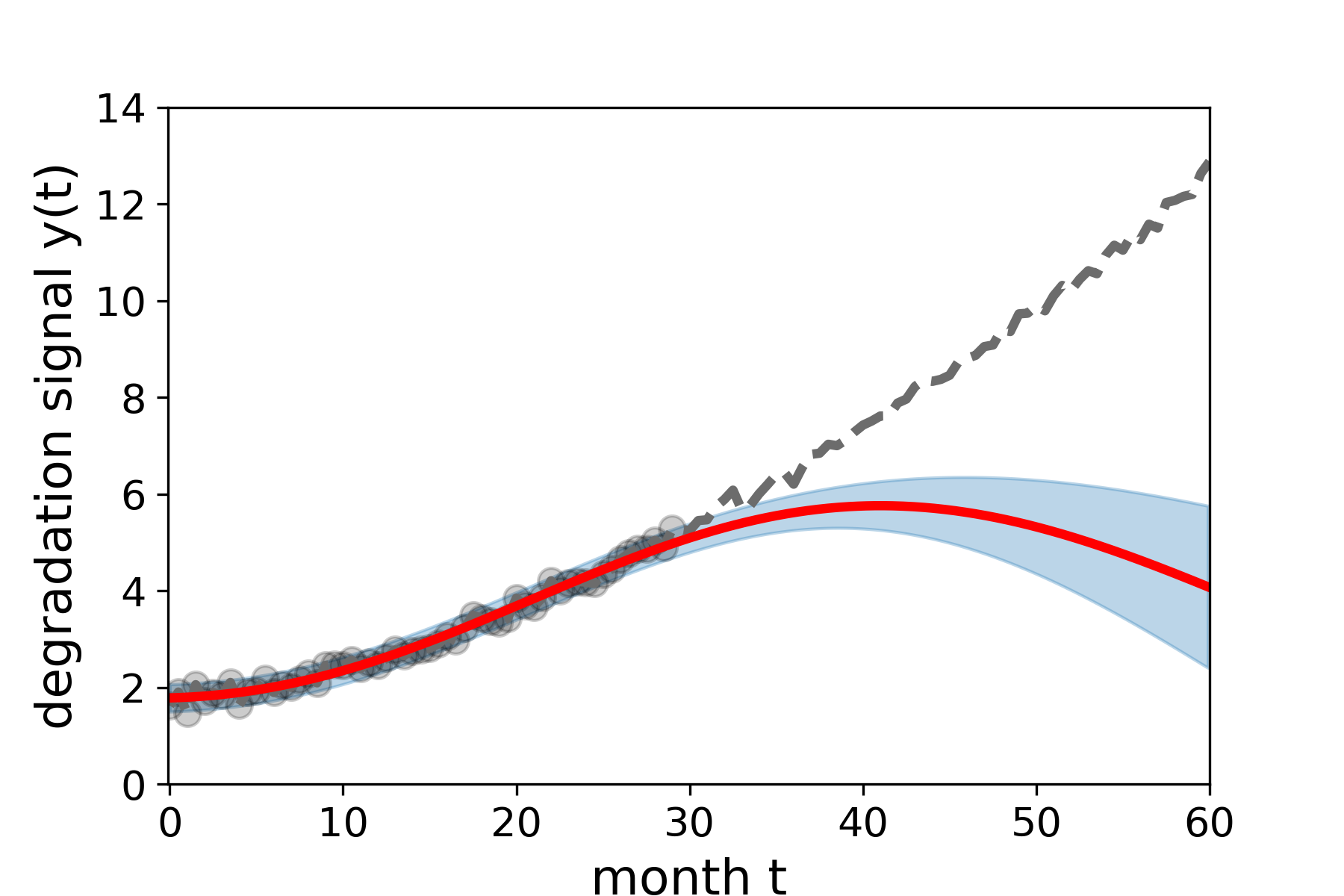}}
\subfigure[LMM-Joint]{\includegraphics[width=0.4\textwidth]{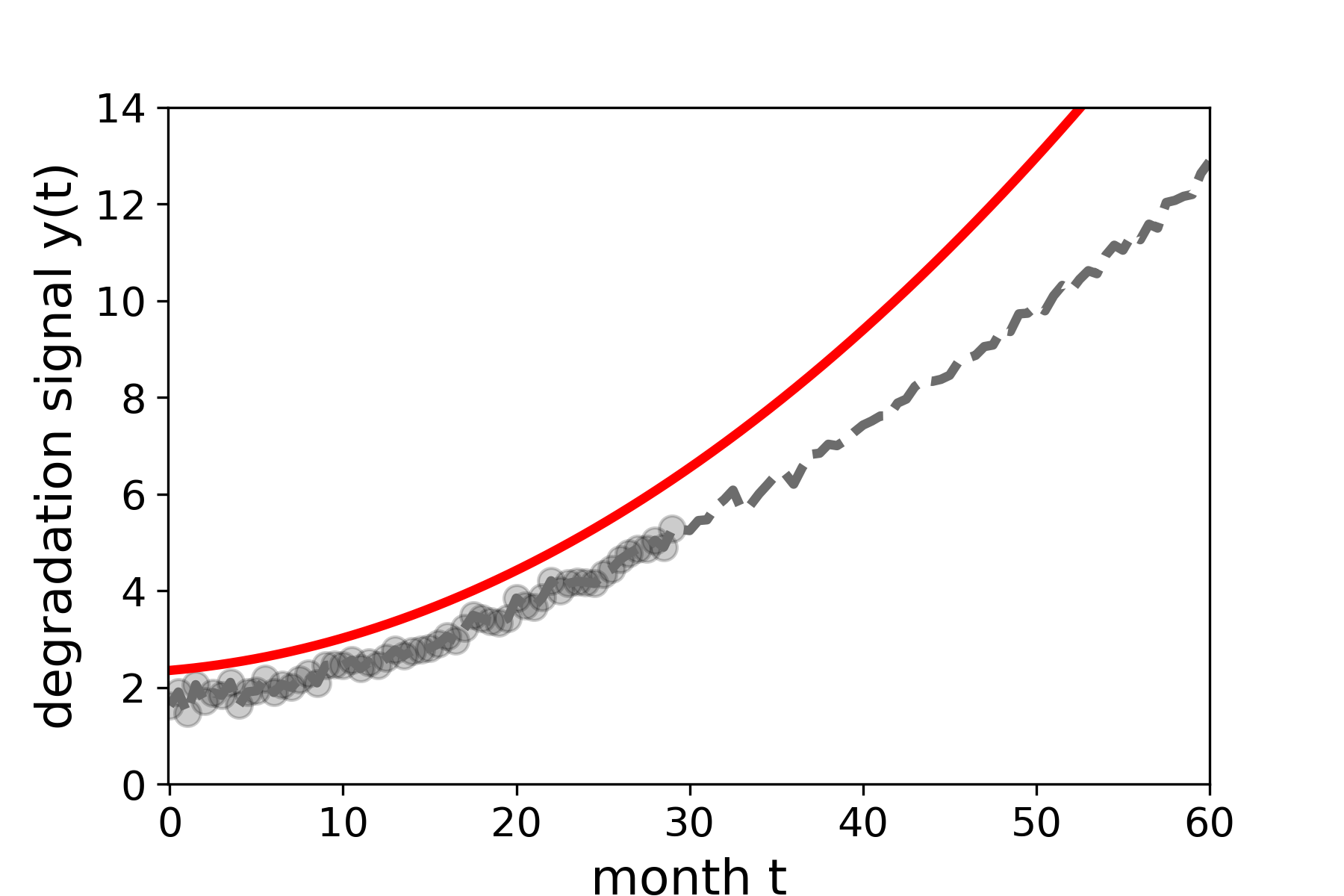}}
\caption{Examples of trajectories for each fitted model in Scenario~I (Note: gray dots represent observations and the rightmost end of dots are prediction times; gray dotted lines represent the true underlying degradation signals; red solid lines represent fitted values using each model; shaded areas represent 95\% confidence bands for each model except for LMM-Joint whose confidence band is omitted).}
\label{fig:simulation_scenario1}
\end{figure}

\begin{figure}[h]
\centering    
\subfigure[Fed-Joint]{\includegraphics[width=0.4\textwidth]{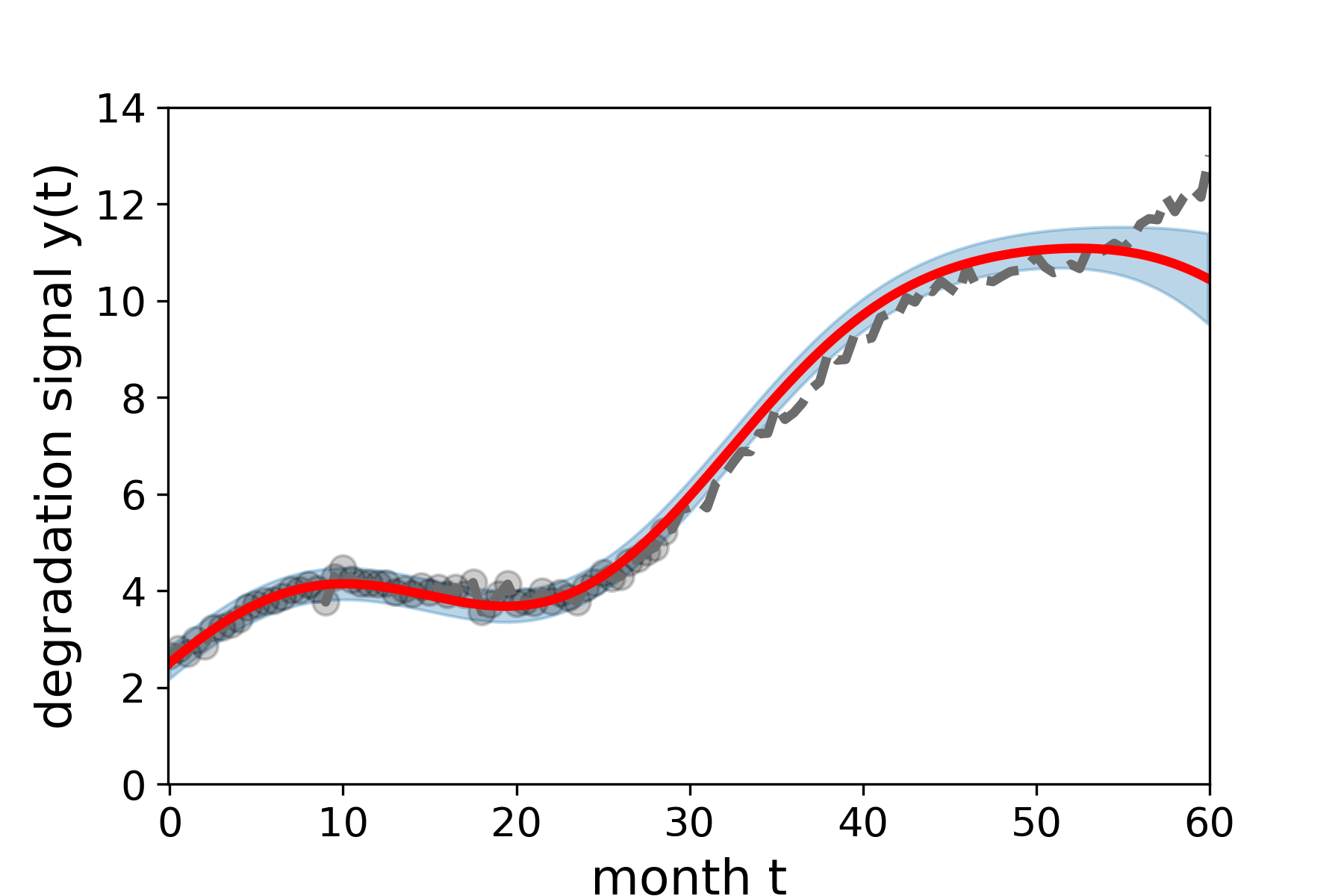}}
\subfigure[Cen-Joint]{\includegraphics[width=0.4\textwidth]{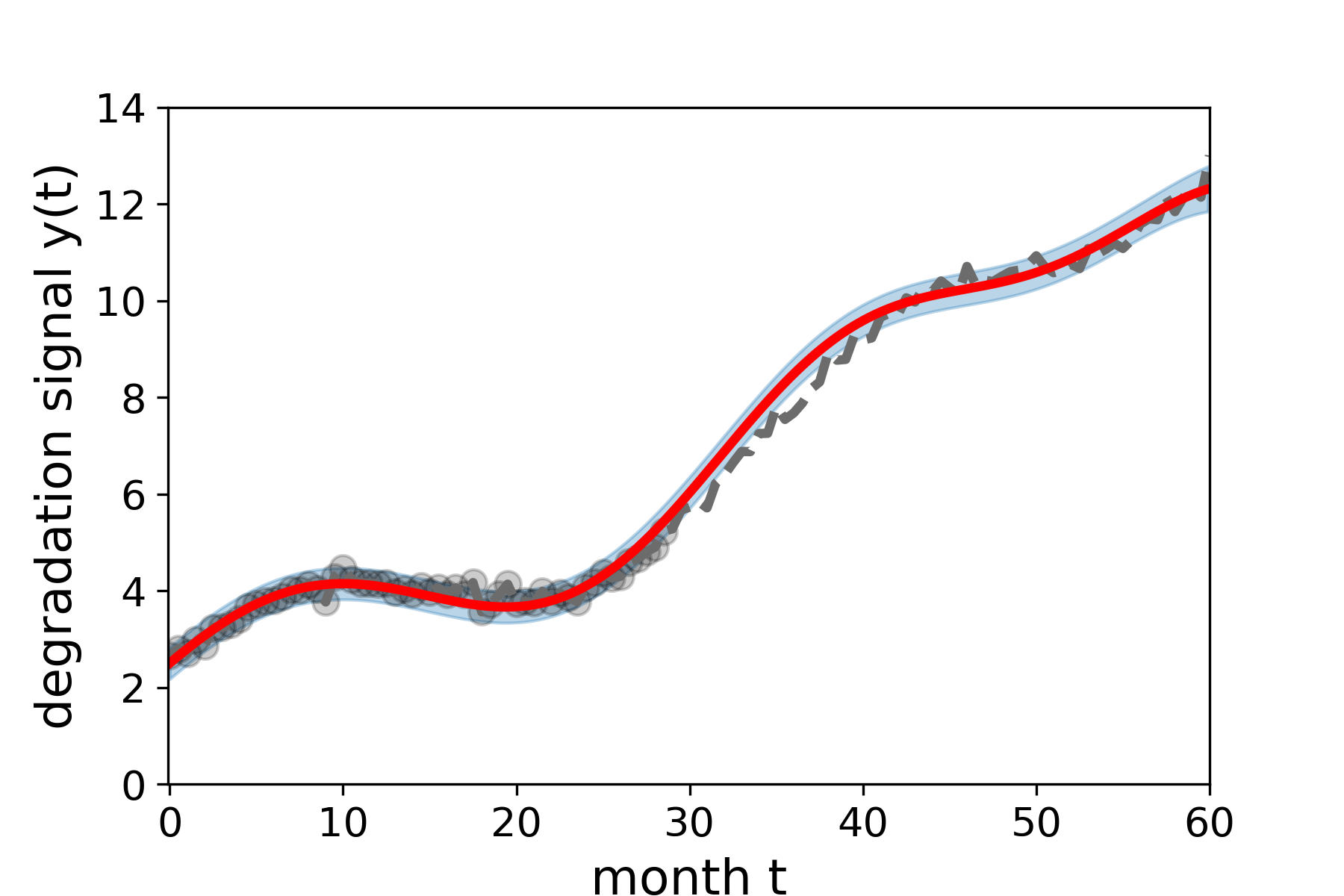}}
\subfigure[Ind-Joint]{\includegraphics[width=0.4\textwidth]{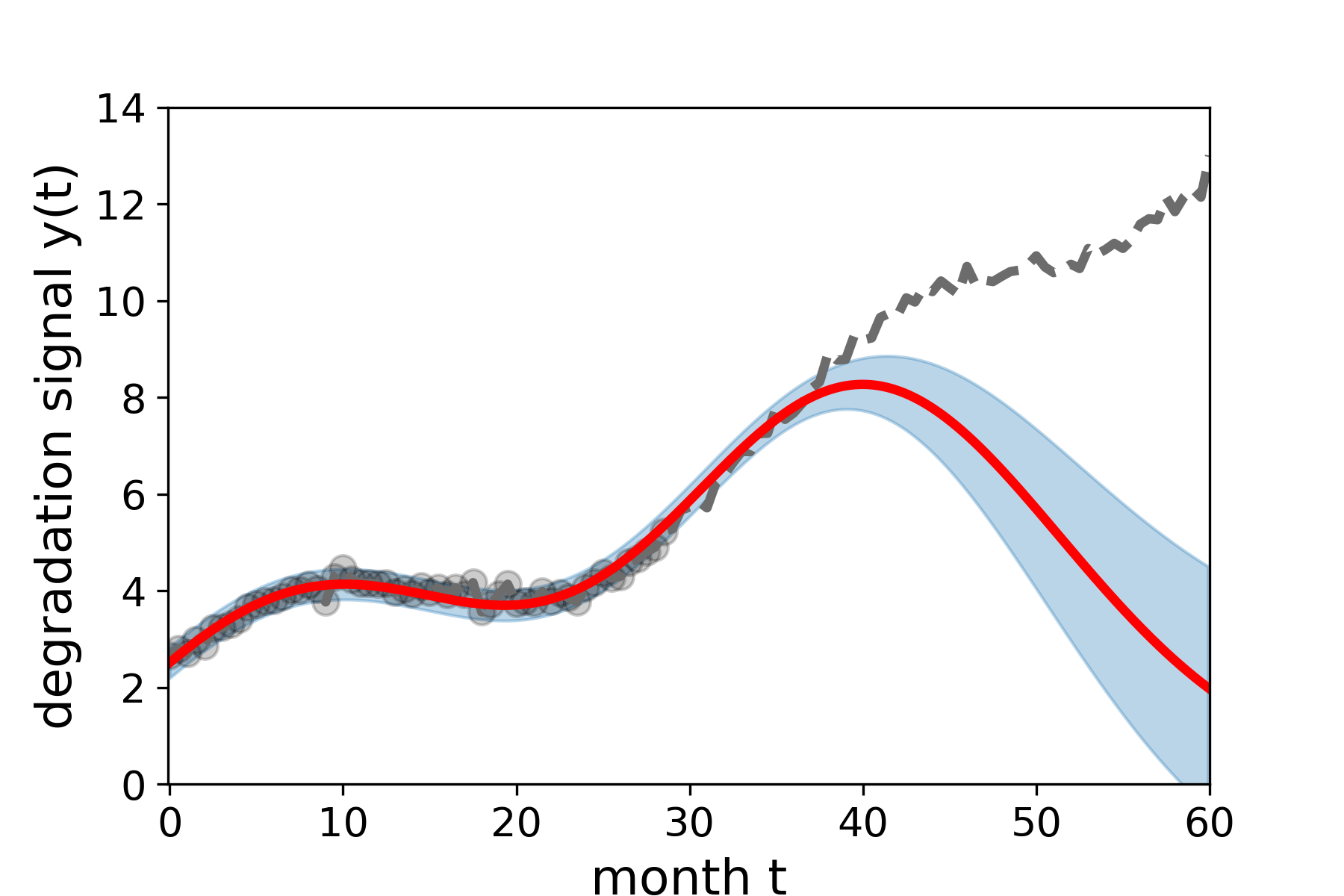}}
\subfigure[LMM-Joint]{\includegraphics[width=0.4\textwidth]{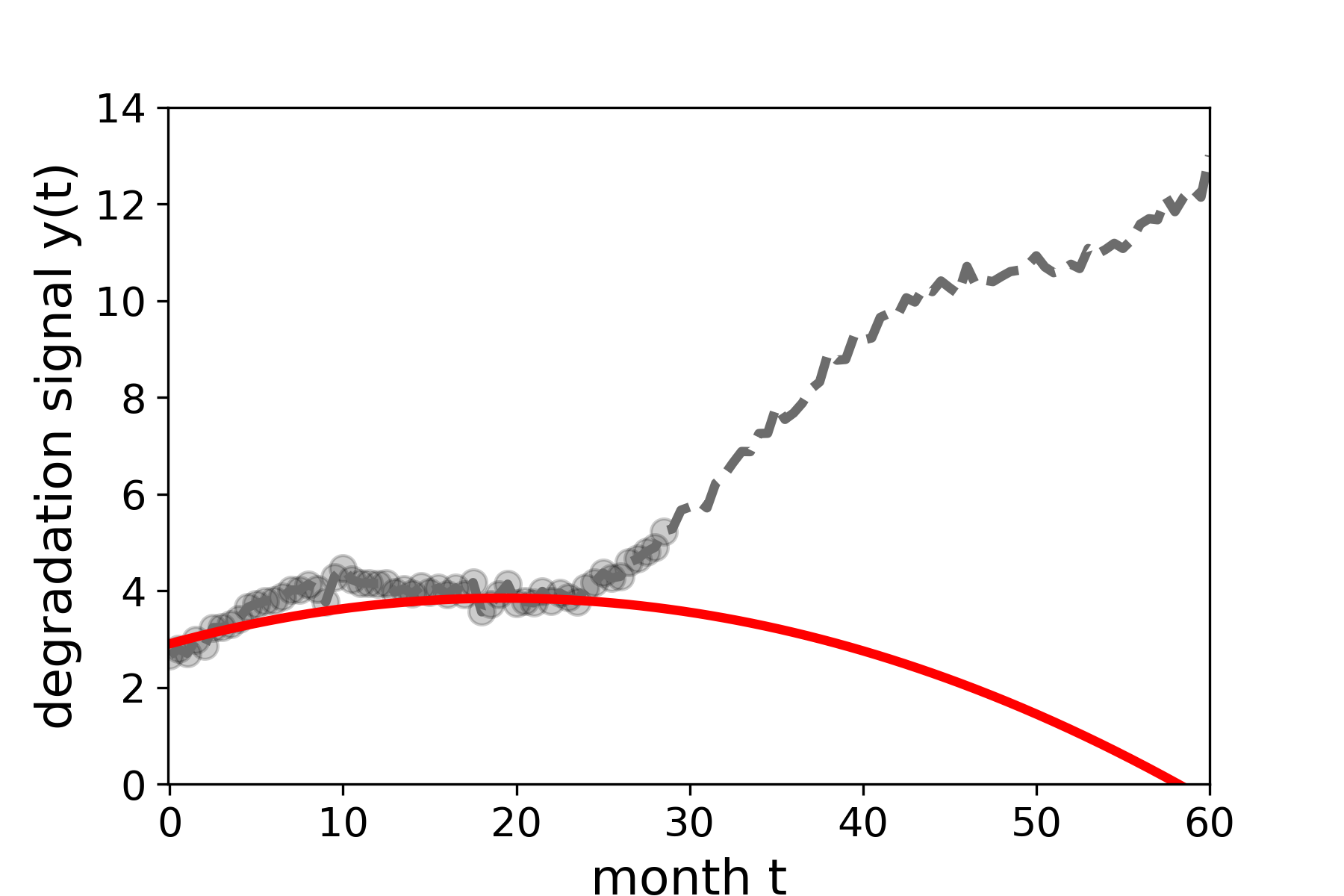}}
\caption{Examples of trajectories for each fitted model in Scenario~II (Note: gray dots represent observations and the rightmost end of dots are prediction times; gray dotted lines represent the true underlying degradation signals; red solid lines represent fitted values using each model; shaded areas represent 95\% confidence bands for each model except for LMM-Joint whose confidence band is omitted).}
\label{fig:simulation_scenario2}
\end{figure}

Specifically, Table~\ref{tb:mae_mrl} shows the comparison results of each method in terms of $MAE_{mrl}$. Notably, Fed-Joint achieves better predictive performance than LMM-Joint in both scenarios. Particularly in Scenario~II, which deals with highly nonlinear signals, Fed-Joint significantly outperforms LMM-Joint. This is because, although LMM-Joint can manage to model the relatively easy-to-fit signals, as shown in Figure~\ref{fig:simulation_scenario1}(d), it fails to model the highly nonlinear signals, as depicted in Figure~\ref{fig:simulation_scenario2}(d). Another reason for its inferiority is that LMM-Joint cannot leverage any knowledge from the units at other sites. In contrast, Fed-Joint fits well to both the polynomial growing pattern in Scenario~I and highly nonlinear pattern in Scenario~II, as shown in Figure~\ref{fig:simulation_scenario1}(a) and Figure~\ref{fig:simulation_scenario2}(a), respectively, while utilizing transferred knowledge. In other words, LMM-Joint fails to predict the future with signals that are not of polynomial form, whereas Fed-Joint accurately predicts future evolution, even with complex signal forms, due to its modeling flexibility and effective transfer learning capabilities.

\begin{table}[H]
\footnotesize
\caption{Comparison of $MAE_{mrl}$ from 20 experiments in the simulation study (Note: the values inside parentheses are standard deviations).}
\label{tb:mae_mrl}
\centering
\scriptsize
\begin{tabular}{@{}c|c|c|ccc|ccc@{}}
\hline\hline
& & & \multicolumn{3}{c}{Scenario~I} \vline & \multicolumn{3}{c}{Scenario~II} \\
\cline{4-9}
\# Sites & \# Units & Method &
$\alpha = 0.3$ & $\alpha = 0.5$ & $\alpha = 0.7$ & $\alpha = 0.3$ & $\alpha = 0.5$ & $\alpha = 0.7$ \\
\hline
3 & 20 & Fed-Joint & 6.93 (1.69) & 6.48 (1.20) & 6.50 (2.17) & 10.11 (3.72) & 7.51 (2.12) & 7.25 (1.92) \\
 &  & Cen-Joint & 6.78 (1.16) & 6.48 (1.12) & 6.51 (1.60) & 7.55 (2.49) & 7.34 (3.62) & 9.62 (10.77) \\
 &  & Ind-Joint & 69.21 (18.42) & 33.5 (7.57) & 12.25 (16.41) & 183.70 (6.87) & 95.36 (28.26) & 37.92 (16.46) \\
 &  & LMM-Joint & 7.13 (1.10) & 6.79 (0.98) & 6.17 (1.04) & 232.95 (9.93) & 231.02 (7.80) & 73.20 (43.38) \\
 \hline
3 & 50 & Fed-Joint & 7.06 (0.60) & 6.61 (0.55) & 5.83 (0.56) & 8.29 (1.60) & 7.09 (1.93) & 6.47 (1.22) \\
 &  & Cen-Joint & 7.22 (0.73) & 10.18 (15.07) & 5.81 (0.57) & 7.22 (1.24) & 6.89 (1.81) & 9.02 (8.91) \\
 &  & Ind-Joint & 90.89 (6.13) & 30.73 (16.87) & 9.55 (13.65) & 185.12 (9.49) & 70.14 (29.84) & 23.8 (11.56) \\
 &  & LMM-Joint & 7.08 (0.55) & 6.82 (0.54) & 6.17 (0.57) & 232.79 (8.51) & 232.91 (4.93) & 96.11 (26.83) \\
 \hline
5 & 20 & Fed-Joint & 6.68 (1.04) & 6.45 (1.05) & 5.73 (1.10) & 8.73 (2.29) & 6.50 (1.48) & 6.15 (1.49) \\
 &  & Cen-Joint & 7.02 (1.05) & 6.49 (1.09) & 5.78 (1.10) & 7.16 (1.55) & 8.22 (8.59) & 7.95 (9.83) \\
 &  & Ind-Joint & 69.21 (18.42) & 33.50 (7.57) & 12.25 (16.41) & 183.70 (6.87) & 95.36 (28.26) & 37.92 (16.46) \\
 &  & LMM-Joint & 7.13 (1.10) & 6.79 (0.98) & 6.17 (1.04) & 232.95 (9.93) & 231.02 (7.80) & 73.20 (43.38) \\
\hline\hline
\end{tabular}
\end{table}

Next, Fed-Joint demonstrates comparable performance to Cen-Joint in both scenarios. This highlights Fed-Joint's capability when data are not shared across sites, potentially reducing the computing and storage needs at the central server without sacrificing prediction accuracy. It can be observed that, even though insufficient data are available, Fed-Joint nearly matches the performance of Cen-Joint, as shown in Figure~\ref{fig:simulation_scenario1}(a) versus Figure~\ref{fig:simulation_scenario1}(b), and Figure~\ref{fig:simulation_scenario2}(a) versus Figure~\ref{fig:simulation_scenario2}(b), thanks to its knowledge transfer capability across sites. Additionally, the predictive performance of Fed-Joint is significantly better than that of Ind-Joint, because our approach collaboratively learns the models using FL for both MGP and CoxPH models, whereas Ind-Joint only learns from its own site's data, not from other sites. This causes Ind-Joint to fail in GP modeling for the future extrapolated area, as depicted in Figure~\ref{fig:simulation_scenario1}(c) and Figure~\ref{fig:simulation_scenario2}(c).

Further, in Tables~\ref{tb:mae_prob1} and \ref{tb:mae_prob2}, they demonstrate how well the compared methods predict the failure probabilities within the future time interval $\Delta t$ at each prediction time $t^*$ in Scenario~I and Scenario~II, respectively. The overall observation is that the prediction accuracy, measured using the error between the true and estimated failure probabilities, i.e., $MAE_{F}$, decreases as $\Delta t$ decreases. It means that better predictions can be made for the near future. This is reasonable because it generally becomes harder to extrapolate the CM signals when the signal to be predicted is farther from the prediction point $t^*$. Aside from that, the general trend in prediction capability follows that of $MAE_{mrl}$, as described above with Table~\ref{tb:mae_mrl}.

\begin{table}[H]
\footnotesize
\caption{Comparison of $MAE_{F}$ from 20 experiments in Scenario~I of the simulation study (Note: the values inside parentheses are standard deviations).}
\label{tb:mae_prob1}
\centering
\resizebox{\textwidth}{!}{\begin{tabular}{@{}c|c|c|ccc|ccc|ccc@{}}
\hline\hline
& & & \multicolumn{3}{c}{$\alpha = 0.3$} \vline & \multicolumn{3}{c}{$\alpha = 0.5$} \vline & \multicolumn{3}{c}{$\alpha = 0.7$} \\
\cline{4-12}
\# Sites & \# Units & Method & $ \Delta t =  12$ & $ \Delta t = 15$ & $ \Delta t = 18$ & $ \Delta t = 12$ & $ \Delta t = 15$ & $ \Delta t = 18$ & $ \Delta t = 12$ & $ \Delta t = 15$ & $ \Delta t = 18$  \\
\hline
3 & 20 & Fed-Joint & 0.02 (0.01) & 0.03 (0.01) & 0.04 (0.01) & 0.04 (0.01) & 0.06 (0.02) & 0.08 (0.03) & 0.10 (0.03) & 0.12 (0.04) & 0.12 (0.04) \\
 &  & Cen-Joint & 0.03 (0.01) & 0.04 (0.01) & 0.05 (0.02) & 0.06 (0.02) & 0.08 (0.03) & 0.11 (0.04) & 0.13 (0.04) & 0.15 (0.05) & 0.16 (0.06) \\
 &  & Ind-Joint & 0.13 (0.01) & 0.15 (0.02) & 0.15 (0.03) & 0.12 (0.02) & 0.12 (0.03) & 0.11 (0.03) & 0.07 (0.04) & 0.07 (0.05) & 0.08 (0.06) \\
 &  & LMM-Joint & 0.04 (0.01) & 0.05 (0.02) & 0.07 (0.02) & 0.06 (0.01) & 0.09 (0.02) & 0.11 (0.02) & 0.12 (0.02) & 0.13 (0.02) & 0.13 (0.03) \\
 \hline
3 & 50 & Fed-Joint & 0.01 (0.00) & 0.02 (0.00) & 0.02 (0.01) & 0.02 (0.01) & 0.03 (0.01) & 0.04 (0.02) & 0.05 (0.02) & 0.06 (0.02) & 0.06 (0.02) \\
 &  & Cen-Joint & 0.01 (0.00) & 0.02 (0.01) & 0.03 (0.01) & 0.02 (0.01) & 0.03 (0.02) & 0.05 (0.02) & 0.05 (0.02) & 0.06 (0.03) & 0.06 (0.03) \\
 &  & Ind-Joint & 0.11 (0.01) & 0.12 (0.01) & 0.10 (0.02) & 0.10 (0.02) & 0.10 (0.02) & 0.09 (0.02) & 0.04 (0.03) & 0.05 (0.04) & 0.06 (0.05) \\
 &  & LMM-Joint & 0.04 (0.01) & 0.05 (0.01) & 0.07 (0.01) & 0.07 (0.01) & 0.09 (0.01) & 0.12 (0.02) & 0.12 (0.01) & 0.14 (0.02) & 0.13 (0.02) \\
 \hline
5 & 20 & Fed-Joint & 0.01 (0.00) & 0.02 (0.00) & 0.02 (0.01) & 0.02 (0.01) & 0.03 (0.01) & 0.04 (0.02) & 0.04 (0.02) & 0.05 (0.03) & 0.05 (0.03) \\
 &  & Cen-Joint & 0.01 (0.01) & 0.02 (0.01) & 0.03 (0.02) & 0.03 (0.02) & 0.04 (0.02) & 0.05 (0.03) & 0.06 (0.03) & 0.07 (0.04) & 0.08 (0.05) \\
 &  & Ind-Joint & 0.13 (0.01) & 0.15 (0.02) & 0.15 (0.03) & 0.12 (0.02) & 0.12 (0.03) & 0.11 (0.03) & 0.07 (0.04) & 0.07 (0.05) & 0.08 (0.06) \\
 &  & LMM-Joint & 0.04 (0.01) & 0.05 (0.02) & 0.07 (0.02) & 0.06 (0.01) & 0.09 (0.02) & 0.11 (0.02) & 0.12 (0.02) & 0.13 (0.02) & 0.13 (0.03) \\
\hline\hline
\end{tabular}}
\end{table}

\begin{table}[H]
\footnotesize
\caption{Comparison of $MAE_{F}$ from 20 experiments in Scenario~II of the simulation study (Note: the values inside parentheses are standard deviations).}
\label{tb:mae_prob2}
\centering
\resizebox{\textwidth}{!}{\begin{tabular}{@{}c|c|c|ccc|ccc|ccc@{}}
\hline\hline
& & & \multicolumn{3}{c}{$\alpha = 0.3$} \vline & \multicolumn{3}{c}{$\alpha = 0.5$} \vline & \multicolumn{3}{c}{$\alpha = 0.7$} \\
\cline{4-12}
\# Sites & \# Units & Method & $ \Delta t =  12$ & $ \Delta t = 15$ & $ \Delta t = 18$ & $ \Delta t = 12$ & $ \Delta t = 15$ & $ \Delta t = 18$ & $ \Delta t = 12$ & $ \Delta t = 15$ & $ \Delta t = 18$  \\
\hline
3 & 20 & Fed-Joint & 0.02 (0.00) & 0.03 (0.01) & 0.04 (0.01) & 0.04 (0.01) & 0.07 (0.01) & 0.10 (0.02) & 0.10 (0.02) & 0.11 (0.02) & 0.11 (0.03) \\
 &  & Cen-Joint & 0.02 (0.00) & 0.03 (0.01) & 0.04 (0.01) & 0.04 (0.01) & 0.07 (0.02) & 0.11 (0.02) & 0.12 (0.04) & 0.14 (0.06) & 0.14 (0.08) \\
 &  & Ind-Joint & 0.05 (0.01) & 0.08 (0.02) & 0.13 (0.02) & 0.08 (0.02) & 0.12 (0.04) & 0.19 (0.07) & 0.13 (0.04) & 0.16 (0.07) & 0.19 (0.08) \\
 &  & LMM-Joint & 0.07 (0.02) & 0.10 (0.02) & 0.14 (0.02) & 0.16 (0.03) & 0.27 (0.04) & 0.41 (0.05) & 0.27 (0.04) & 0.37 (0.06) & 0.44 (0.09) \\
 \hline
3 & 50 & Fed-Joint & 0.02 (0.00) & 0.03 (0.00) & 0.04 (0.01) & 0.04 (0.00) & 0.06 (0.01) & 0.09 (0.01) & 0.09 (0.01) & 0.10 (0.02) & 0.10 (0.02) \\
 &  & Cen-Joint & 0.02 (0.00) & 0.02 (0.00) & 0.04 (0.00) & 0.04 (0.00) & 0.06 (0.01) & 0.09 (0.01) & 0.10 (0.03) & 0.12 (0.05) & 0.12 (0.07) \\
 &  & Ind-Joint & 0.06 (0.02) & 0.08 (0.02) & 0.13 (0.02) & 0.10 (0.02) & 0.11 (0.02) & 0.15 (0.06) & 0.10 (0.02) & 0.12 (0.03) & 0.13 (0.04) \\
 &  & LMM-Joint & 0.07 (0.02) & 0.10 (0.02) & 0.14 (0.03) & 0.16 (0.02) & 0.26 (0.02) & 0.40 (0.03) & 0.30 (0.02) & 0.41 (0.03) & 0.50 (0.05) \\
 \hline
5 & 20 & Fed-Joint & 0.02 (0.00) & 0.03 (0.00) & 0.04 (0.01) & 0.04 (0.01) & 0.06 (0.01) & 0.09 (0.02) & 0.09 (0.02) & 0.10 (0.02) & 0.10 (0.02) \\
 &  & Cen-Joint & 0.02 (0.00) & 0.02 (0.00) & 0.04 (0.01) & 0.04 (0.01) & 0.06 (0.02) & 0.09 (0.02) & 0.10 (0.03) & 0.12 (0.05) & 0.12 (0.07) \\
 &  & Ind-Joint & 0.05 (0.01) & 0.08 (0.02) & 0.13 (0.02) & 0.08 (0.02) & 0.12 (0.04) & 0.19 (0.07) & 0.13 (0.04) & 0.16 (0.07) & 0.19 (0.08) \\
 &  & LMM-Joint & 0.07 (0.02) & 0.10 (0.02) & 0.14 (0.02) & 0.16 (0.03) & 0.27 (0.04) & 0.41 (0.05) & 0.27 (0.04) & 0.37 (0.06) & 0.44 (0.09) \\
\hline
\hline
\end{tabular}}
\end{table}

\section{Case Study: Remaining Useful Life Prediction for Turbofan Engines in Distributed Sites}
\label{sec:case}

In this section, we demonstrate the effectiveness of Fed-Joint through a case study. In Section~\ref{sec:case_description}, we introduce the dataset of multi-channel turbofan engine degradation signals and run-to-failure times. Section~\ref{sec:case_result} discusses the results by comparing the federated joint modeling framework using Fed-Joint with the centralized version that uses Cen-Joint.

\subsection{Data Description}
\label{sec:case_description}

\begin{figure}[h]
\centering    
\includegraphics[width=0.42\textwidth]{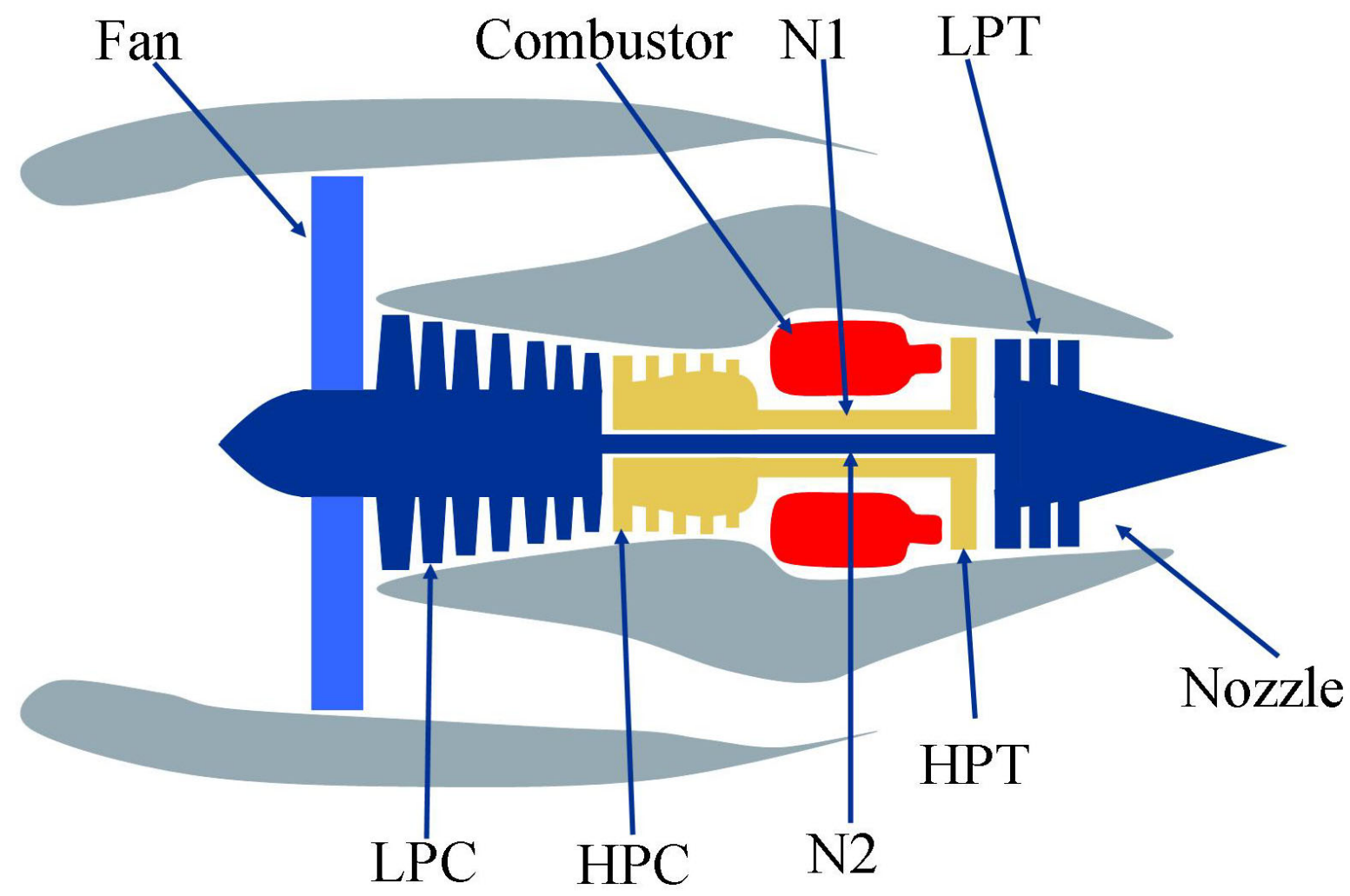}
\caption{Diagram of a commercial turbofan engine used for simulation in C-MAPSS (Note: LPC/HPC stand for low-/high-pressure compressor; LPT/HPT stand for low-/high-pressure turbine)~\citep{frederick2007cmapss}.}
\label{fig:turbofan_diagram}
\end{figure}

The turbofan engine degradation dataset, provided by the National Aeronautics and Space Administration (NASA), includes time-series degradation signals from multiple sensors installed on a fleet of aircraft turbofan engines. Figure~\ref{fig:turbofan_diagram} presents a simplified diagram of the turbofan engine. The dataset is generated by the Commercial Modular Aero-Propulsion System Simulation (C-MAPSS)~\citep{frederick2007cmapss}, a simulation software developed in the MATLAB and Simulink environment. By adjusting input parameters and specifying operating conditions within C-MAPSS, users can simulate the impact of faults and deterioration in various rotating components such as the Fan, Low-Pressure Compressor (LPC), High-Pressure Compressor (HPC), High-Pressure Turbine (HPT), and Low-Pressure Turbine (LPT) of the engines. Simulation is conducted under four different conditions, with the datasets denoted as FD001$, \ldots, $ FD004. For this case study, we use the FD001 dataset that contains a total of 100 signals, assuming that engine degradation results from wear and tear of a single component, specifically the HPC, under constant operating conditions~\citep{saxena2008damage}. 

\begin{figure}[htb!]
\centering    
\subfigure[Sensor~4]{\includegraphics[width=0.4\textwidth]{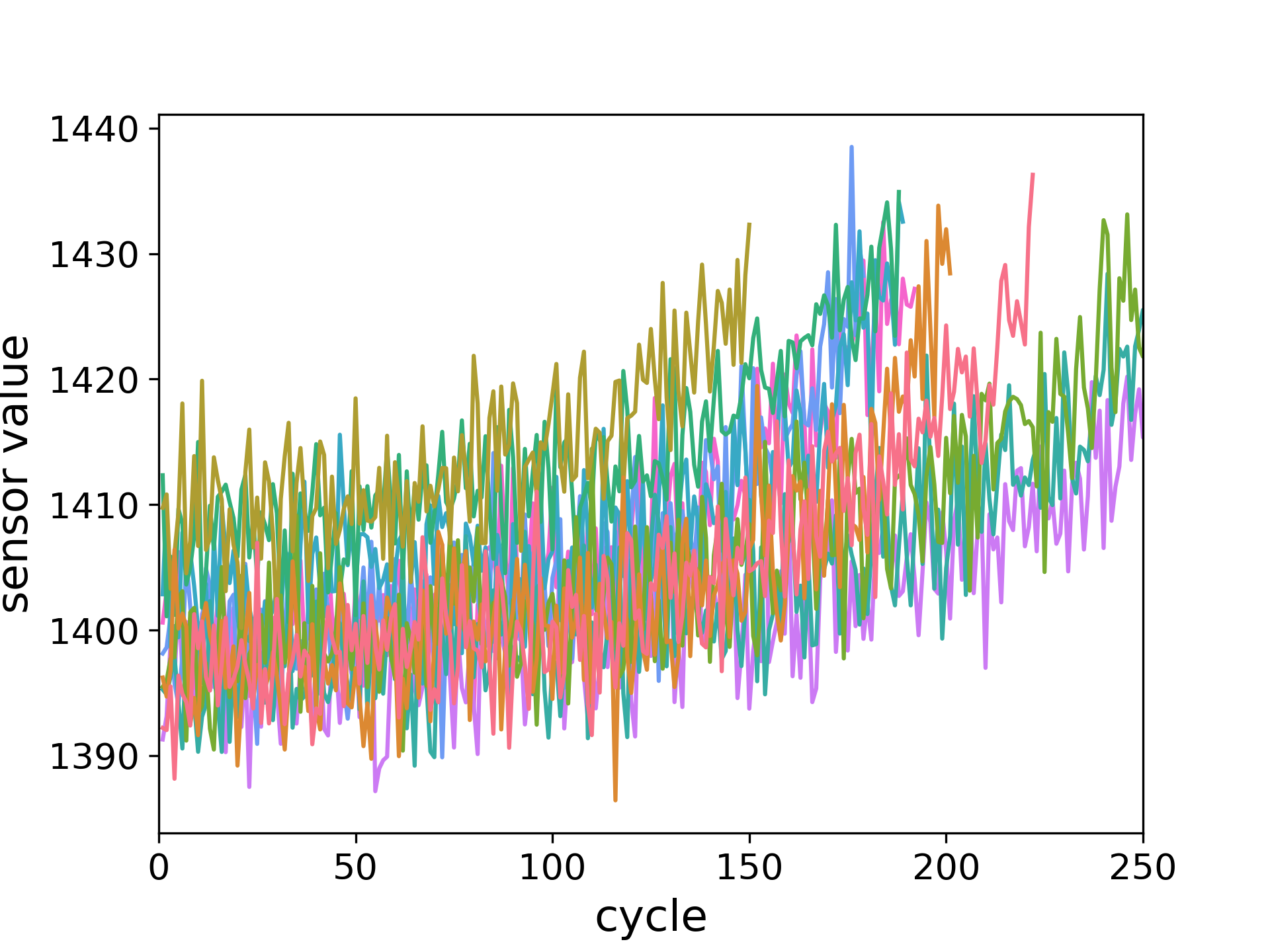}}
\subfigure[Sensor~15]{\includegraphics[width=0.4\textwidth]{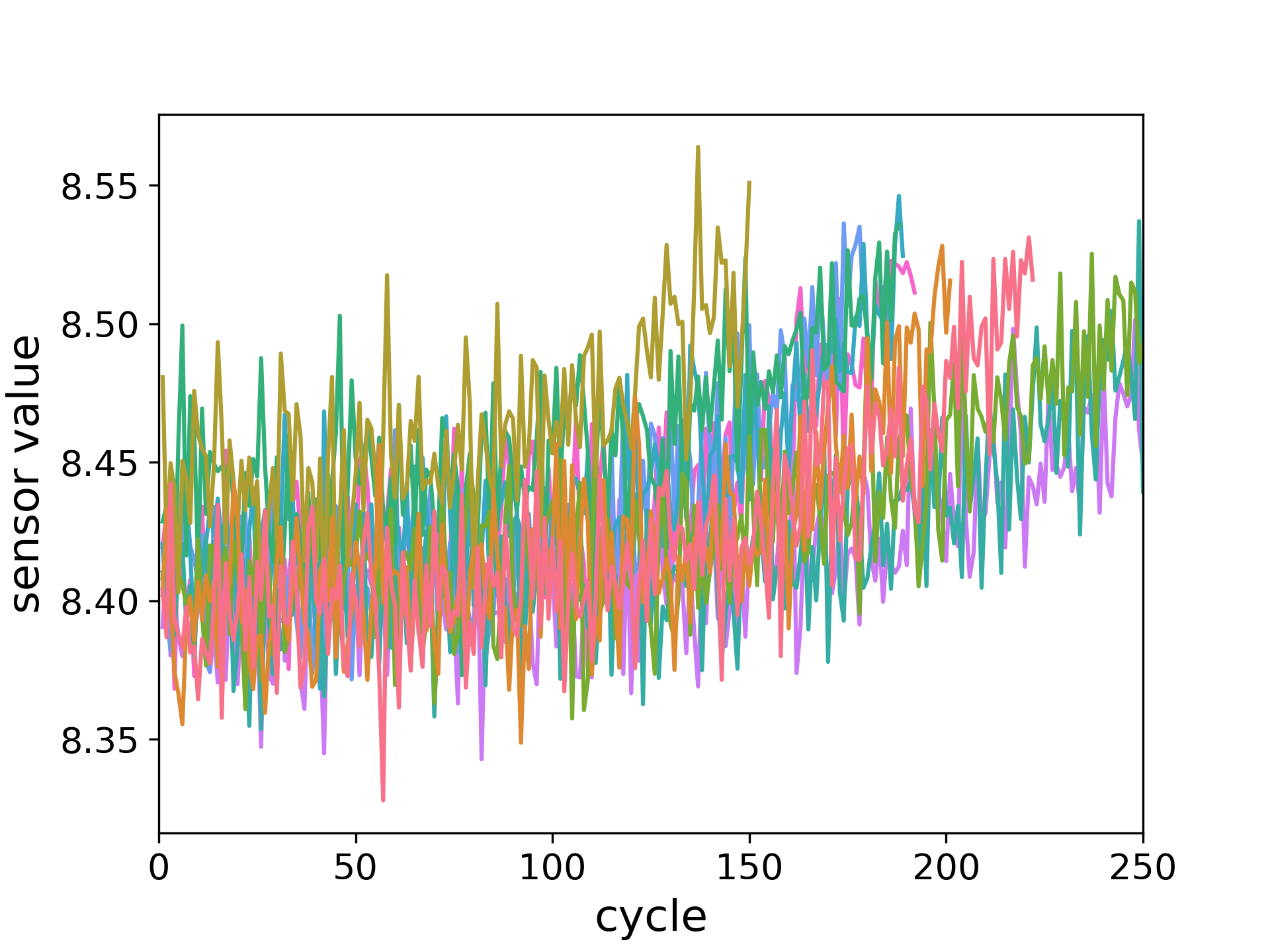}}
\subfigure[Sensor~17]{\includegraphics[width=0.4\textwidth]{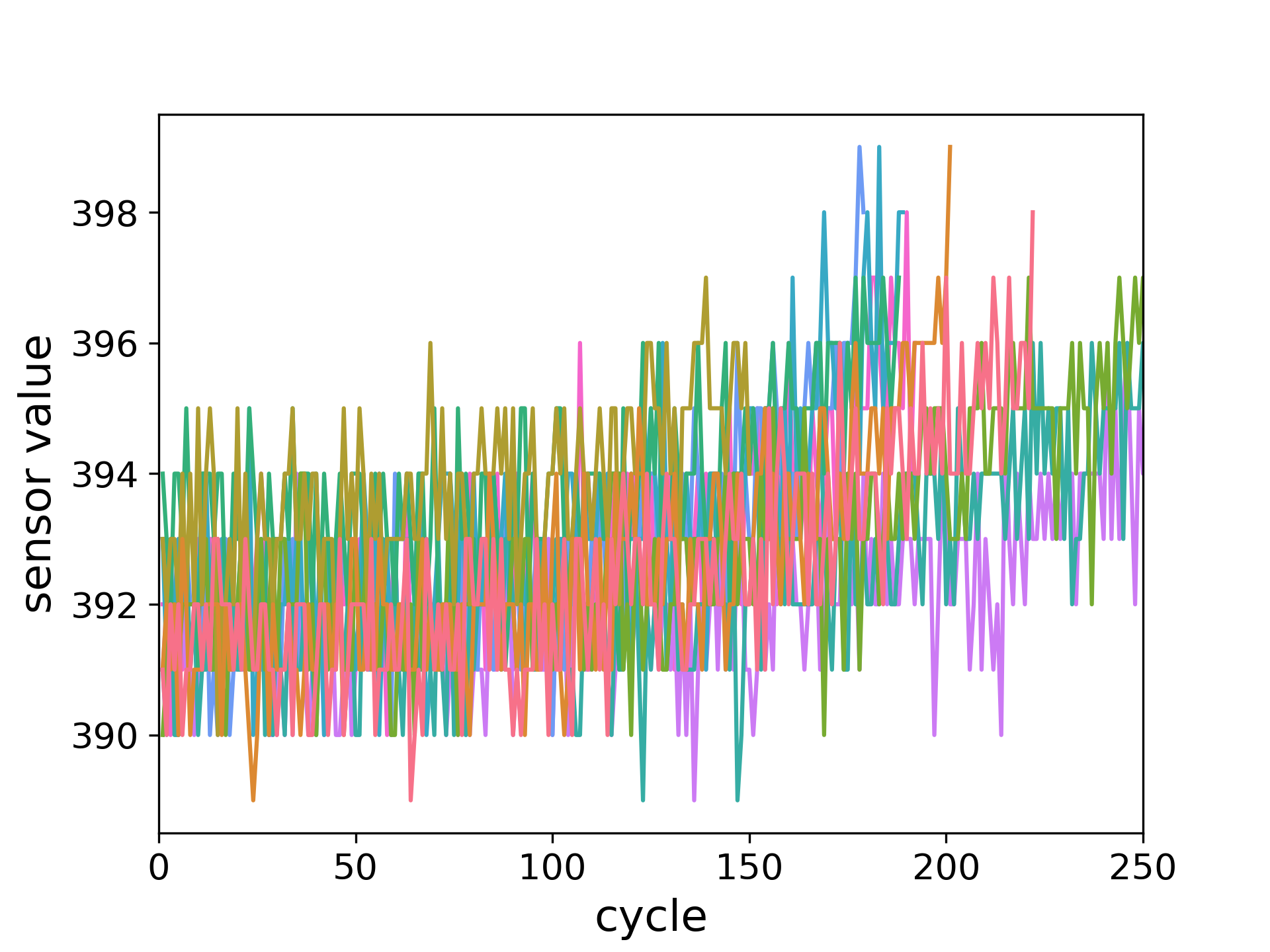}}
\subfigure[Sensor~20]{\includegraphics[width=0.4\textwidth]{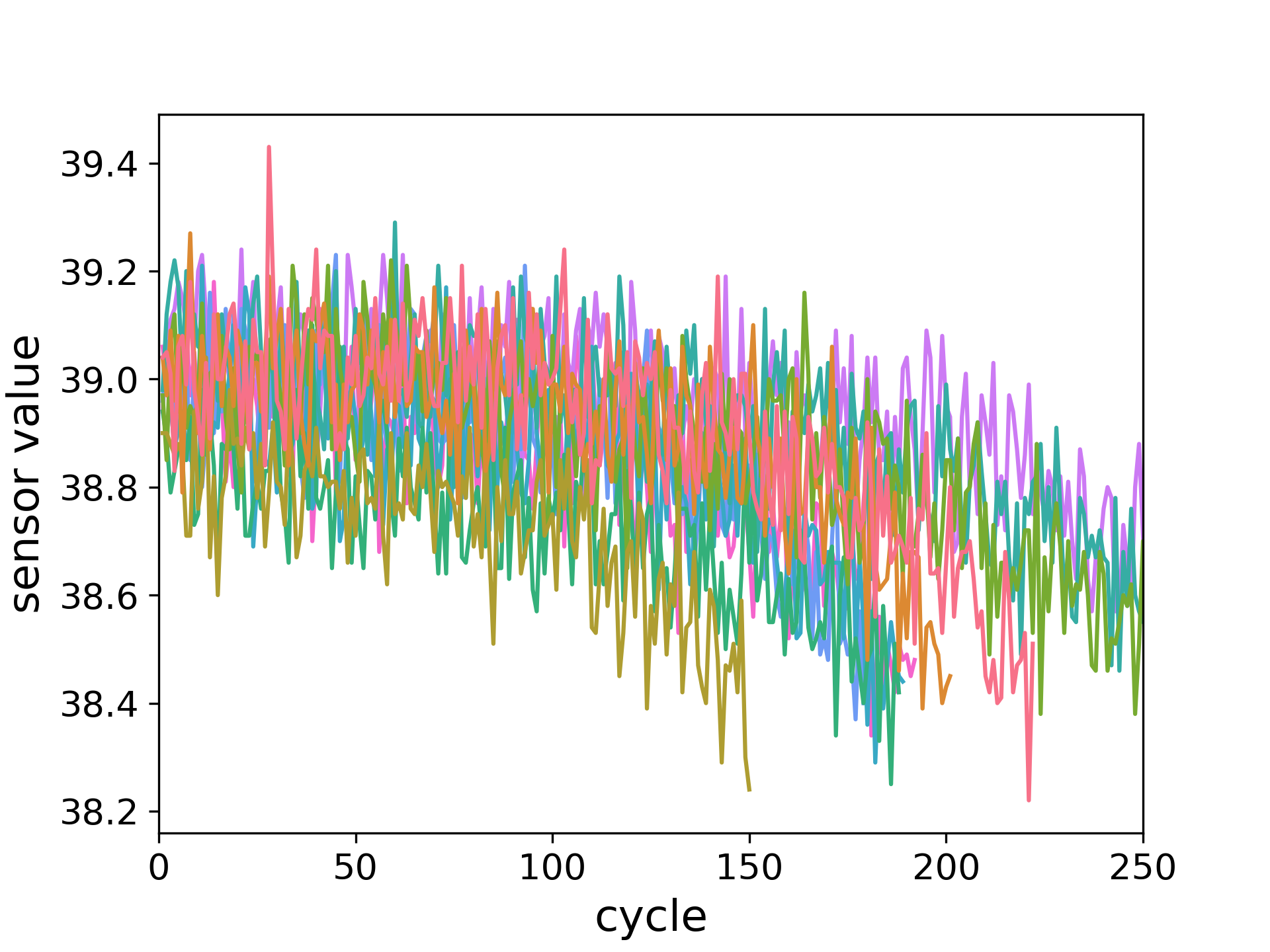}}
\caption{Examples of 10 randomly sampled degradation signals for each sensor.}
\label{fig:degradation_signals_case}
\end{figure}

Based on the analysis in~\cite{fang2017multistream}, 4 out of 21 sensors play an important role in predicting the health status of a component or the entire engine: \textit{total temperature at LPT outlet} (Sensor 4), \textit{bypass ratio} (Sensor 15), \textit{bleed enthalpy} (Sensor 17), and \textit{HPT coolant bleed} (Sensor 20), as indexed in~\cite{saxena2008damage}. Figure~\ref{fig:degradation_signals_case} shows examples of degradation signals from these four sensors. 

For the implementation, we set the threshold for the ending cycle at 250. If the signals are recorded and ended beyond this threshold, they are considered right-censored. If the signals are recorded before the threshold, we consider these times as failure times. Now, we assume there are three sites, and the raw data collected from each site are not shared across sites (see Figure~\ref{fig:dist}). To represent this federated scenario, for each sensor, we randomly sample 20 out of 100 signals (or units) and assign them to the testing site. Among the 80 remaining signals, we randomly sample 40 signals and assign them to two training sites, 20 each, so that we have $(K,M_k) = (3,20)$ for sites $k=1,2,3$. For the units in the testing site, we consider three levels of data incompleteness: 30\%, 50\%, and 70\% (i.e., $\alpha=0.3,0.5,0.7$), as in the simulation studies. Thus, the prediction time $t^*$ for each unit corresponds to 30\%, 50\%, and 70\% of the ending cycle in the testing site, that is, $t^* = \lceil \alpha \times \text{(ending cycle)} \rceil$. Using a similar procedure as above, we create 20 different training and test sets for the repeated experiments for each sensor. For ease of implementation, we use the exponential density function for the baseline hazard rate function $h_0(t)$, i.e., $h_0(t) = \lambda$. We assume that the time-invariant covariate $\bm{w}_{k,m} = 0$, reflecting the possibility that the degradation signals may not be heterogeneous in this application.

\subsection{Results and Analysis}
\label{sec:case_result}

Using the preprocessed datasets for the four informative sensors, we evaluate the predictive performance of Fed-Joint and Cen-Joint by calculating the averaged predictive performance using $MAE_{mrl}$ and $MAE_{F}$ for each sensor, as summarized in Table~\ref{tb:case_mrl} and Table~\ref{tb:case_prob}, respectively. Since it is evident that our proposed Fed-Joint model achieves much higher prediction accuracy than other alternatives, we only compare Fed-Joint with Cen-Joint in this case study.

Clearly, in Tables~\ref{tb:case_mrl} and~\ref{tb:case_prob}, we observe that prediction accuracy increases as $\alpha$ increases for all sensors. Similar to the simulation studies, this is because the accuracy of the fitted degradation signals improves with more observations. Further, in Table~\ref{tb:case_prob}, we observe that $MAE_{F}$ decreases as $\Delta t$ decreases, meaning that better predictions can be made for the near future.

From Tables~\ref{tb:case_mrl} and~\ref{tb:case_prob}, we also observe that the prediction performance of Fed-Joint is nearly comparable to that of Cen-Joint. This indicates that even if the data are not shared across sites and each site does not have enough units for model training, Fed-Joint successfully borrows knowledge from other units and sites by building the joint model of nonlinear degradation signals and time-to-failure data by leveraging the FL technique. Thus, we claim that when data are not shared across sites and significant costs are associated with data transfer, computing, and storage, we should be inclined to use Fed-Joint rather than Cen-Joint to predict failure times and probabilities for target units.

\begin{table}[htb!]
\footnotesize
\caption{Comparison of $MAE_{mrl}$ from 20 experiments in the case study (Note: the values inside parentheses are standard deviations).}
\label{tb:case_mrl}
\centering
\begin{tabular}{@{}c|c|ccc@{}}
\hline\hline
Sensor & Method & $\alpha = 0.3$ & $\alpha = 0.5$ & $\alpha = 0.7$ \\
\hline
4 & Fed-Joint & 30.71 (4.89) & 24.82 (3.12) & 21.16 (3.84) \\
 & Cen-Joint & 32.65 (6.62) & 26.35 (4.53) & 20.56 (3.48) \\
 \hline
15 & Fed-Joint & 30.72 (3.56) & 24.58 (3.15) & 20.50 (3.03) \\
 & Cen-Joint & 32.13 (4.61) & 24.89 (3.07) & 20.08 (2.96) \\
 \hline
17 & Fed-Joint & 32.31 (6.27) & 24.86 (4.32) & 20.81 (2.74) \\
 & Cen-Joint & 33.54 (7.05) & 26.05 (4.68) & 20.48 (2.67) \\
 \hline
20 & Fed-Joint & 32.09 (4.50) & 24.65 (3.22) & 19.27 (2.52) \\
 & Cen-Joint & 32.23 (4.30) & 24.73 (3.14) & 19.18 (2.58) \\
\hline\hline
\end{tabular}
\end{table}

\begin{table}[htb!]
\footnotesize
\caption{Comparison of $MAE_{F}$ from 20 experiments in the case study (Note: the values inside parentheses are standard deviations).}
\label{tb:case_prob}
\centering
\resizebox{\textwidth}{!}{\begin{tabular}{@{}c|c|ccc|ccc|ccc@{}}
\hline\hline
& & \multicolumn{3}{c}{$\alpha = 0.3$} \vline & \multicolumn{3}{c}{$\alpha = 0.5$} \vline & \multicolumn{3}{c}{$\alpha = 0.7$} \\
\cline{3-11}
Sensor & Method & $ \Delta t =  50$ & $ \Delta t = 70$ & $ \Delta t = 90$ & $ \Delta t = 50$ & $ \Delta t = 70$ & $ \Delta t = 90$ & $ \Delta t = 50$ & $ \Delta t = 70$ & $ \Delta t = 90$  \\
\hline
4 & Fed-Joint & 0.184 (0.010) & 0.248 (0.012) & 0.307 (0.015) & 0.184 (0.010) & 0.248 (0.012) & 0.307 (0.015) & 0.184 (0.010) & 0.248 (0.012) & 0.307 (0.015) \\
 & Cen-Joint & 0.188 (0.012) & 0.253 (0.015) & 0.313 (0.018) & 0.189 (0.013) & 0.255 (0.017) & 0.315 (0.020) & 0.190 (0.013) & 0.255 (0.017) & 0.315 (0.020) \\
 \hline
15 & Fed-Joint & 0.187 (0.010) & 0.252 (0.012) & 0.311 (0.015) & 0.188 (0.011) & 0.253 (0.014) & 0.313 (0.017) & 0.189 (0.013) & 0.255 (0.017) & 0.315 (0.020) \\
 & Cen-Joint & 0.190 (0.011) & 0.255 (0.014) & 0.316 (0.016) & 0.190 (0.011) & 0.255 (0.014) & 0.316 (0.016) & 0.190 (0.011) & 0.255 (0.014) & 0.316 (0.016) \\
 \hline
17 & Fed-Joint & 0.189 (0.018) & 0.254 (0.022) & 0.314 (0.026) & 0.186 (0.013) & 0.250 (0.017) & 0.309 (0.020) & 0.187 (0.013) & 0.251 (0.017) & 0.311 (0.020) \\
 & Cen-Joint & 0.190 (0.014) & 0.255 (0.019) & 0.315 (0.022) & 0.190 (0.014) & 0.255 (0.019) & 0.315 (0.022) & 0.190 (0.014) & 0.255 (0.019) & 0.315 (0.022) \\
 \hline
20 & Fed-Joint & 0.188 (0.011) & 0.253 (0.014) & 0.312 (0.017) & 0.188 (0.011) & 0.253 (0.014) & 0.312 (0.017) & 0.188 (0.011) & 0.253 (0.014) & 0.312 (0.017) \\
 & Cen-Joint & 0.189 (0.011) & 0.255 (0.014) & 0.315 (0.017) & 0.189 (0.011) & 0.255 (0.014) & 0.315 (0.017) & 0.189 (0.011) & 0.254 (0.014) & 0.314 (0.017) \\
\hline\hline
\end{tabular}}
\end{table}

\section{Conclusion}
\label{sec:conclusion}

We present a new prognostic framework that employs joint modeling of degradation signals and time-to-failure data for RUL prediction when data cannot be shared across sites, such as companies, factories, and production lines, due to confidentiality issues and a lack of storage and computing capabilities. One important aspect is that the form of the signals is not known \textit{a priori}, and they behave in a highly nonlinear fashion. To address this, we adopt FL for both degradation signal and survival model training. Both simulation and real-world case studies are conducted to demonstrate the effectiveness of Fed-Joint over other existing alternatives. In the future, we plan to extend our framework to more general settings such as handling multi-source \citep{yue2024federated1} heterogeneous degradation data, incorporating domain adaptation techniques for meta- and transfer learning, and integrating adaptive personalization strategies to improve predictive performance for individual sites. Additionally, we aim to explore privacy-preserving techniques, such as differential privacy, to further enhance data security in collaborative prognostics. Another possible research direction is to apply MGP with FL to the diagnostic task of identifying the root cause of product quality defects~\citep{Jeong2021,jeong2024tensor}, especially when the multi-channel signals are highly nonlinear and raw data are not shared across different sites.


\section*{Acknowledgement} 
\noindent This work was supported in part by faculty startup at Northeastern University and University of Virginia.

\bibliography{bibliography.bib}

\end{document}